\documentclass{article}
\pdfoutput=1
\usepackage{microtype}
\usepackage{graphicx}
\usepackage{subfigure}
\usepackage{booktabs} 

\usepackage{amsmath}
\usepackage{relsize}
\usepackage{amsmath}
\usepackage{amsfonts}
\usepackage{amsthm}
\usepackage{comment}
\newtheorem{theorem}{Theorem}
\newtheorem{lemma}{Lemma}
\theoremstyle{definition}

\DeclareMathOperator*{\argmax}{arg\,max}
\DeclareMathOperator*{\argmin}{arg\,min}
\usepackage{array}

\usepackage{amsmath}               
{
	\theoremstyle{plain}
	\newtheorem{assumption}{Assumption}
}

\usepackage{dsfont}
\usepackage{enumerate}
\usepackage[shortlabels]{enumitem}
\usepackage[font=small,skip=0pt]{caption}

\usepackage{hyperref}

\newcommand{\torcs}{\textsc{Torcs}}

\usepackage[accepted]{icml2019}

\icmltitlerunning{Control Regularization for Reduced Variance Reinforcement Learning}

\begin{document}

\twocolumn[
\icmltitle{Control Regularization for Reduced Variance Reinforcement Learning}




\begin{icmlauthorlist}
\icmlauthor{Richard Cheng}{ca}
\icmlauthor{Abhinav Verma}{ri}
\icmlauthor{G\'abor Orosz}{mi}
\icmlauthor{Swarat Chaudhuri}{ri}
\icmlauthor{Yisong Yue}{ca}
\icmlauthor{Joel W. Burdick}{ca}
\end{icmlauthorlist}

\icmlaffiliation{ca}{California Institute of Technology, Pasadena, CA}
\icmlaffiliation{mi}{University of Michigan, Ann Arbor, MI}
\icmlaffiliation{ri}{Rice University, Houston, TX}

\icmlcorrespondingauthor{Richard Cheng}{rcheng@caltech.edu}

\icmlkeywords{Machine Learning, ICML}

\vskip 0.3in
]



\printAffiliationsAndNotice{}  

\begin{abstract}
Dealing with high variance is a significant challenge in model-free reinforcement learning (RL). Existing methods are unreliable, exhibiting high variance in performance from run to run using different initializations/seeds. Focusing on problems arising in continuous control, we propose a functional regularization approach to augmenting model-free RL. In particular, we regularize the behavior of the deep policy to be similar to a policy prior, i.e., we regularize in function space.  We show that functional regularization yields a bias-variance trade-off, and propose an adaptive tuning strategy to optimize this trade-off. When the policy prior has control-theoretic stability guarantees, we further show that this regularization approximately preserves those stability guarantees throughout learning. We validate our approach empirically on a range of settings, and demonstrate significantly reduced variance, guaranteed dynamic stability, and more efficient learning than deep RL alone.
\end{abstract}

\section{Introduction}
\label{sec:intro}

Reinforcement learning (RL) focuses on finding an agent's policy (i.e. controller) that maximizes long-term accumulated reward. This is done by the agent repeatedly observing its state, taking an action (according to a policy), and receiving a reward.  Over time the agent modifies its policy to maximize its long-term reward. Amongst other applications, this method has been successfully applied to control tasks \cite{Lillicrap2015,Schulman2015,Ghosh2018}, learning to stabilize complex robots. 

In this paper, we focus particularly on policy gradient (PG) RL algorithms, which have become popular in solving continuous control tasks \cite{Duan2016}. Since PG algorithms focus on maximizing the long-term reward through trial and error, they can learn to control complex tasks without a prior model of the system. This comes at the cost of slow, high variance, learning -- complex tasks can take millions of iterations to learn. More importantly, variation between learning runs can be very high, meaning some runs of an RL algorithm succeed while others fail depending on randomness in initialization and sampling. Several studies have noted this high variability in learning as a significant hurdle for the application of RL, since learning becomes unreliable \cite{Henderson2017,Arulkumaran2017,Recht2018}. All policy gradient algorithms face the same issue. 

We can alleviate the aforementioned issues by introducing a control-theoretic prior into the learning process using functional regularization. Theories and procedures exist to design stable controllers for the vast majority of real-world physical systems (from humanoid robots to robotic grasping to smart power grids). However, conventional controllers for complex systems can be highly suboptimal and/or require great effort in system modeling and controller design. It would be ideal then to leverage \textit{simple}, suboptimal controllers in RL to reliably learn high-performance policies.

In this work, we propose a policy gradient algorithm, CORE-RL (COntrol REgularized Reinforcement Learning), that utilizes a functional regularizer around a, typically suboptimal, control prior (i.e. a controller designed from any prior knowledge of the system). We show that this approach significantly lowers variance in the policy updates, and leads to higher performance policies when compared to both the baseline RL algorithm and the control prior. In addition, we prove that our policy can maintain control-theoretic stability guarantees throughout the learning process. Finally, we empirically validate our approach using three benchmarks: a car-following task with real driving data, the TORCS racecar simulator, and a simulated cartpole problem. In summary, the main contributions of this paper are as follows:
\begin{itemize}[topsep=0pt, leftmargin=*]
	\itemsep-0.2em
	\item We introduce functional regularization using a control prior, and prove that this significantly reduces variance during learning at the cost of potentially increasing bias.
	\item We provide control-theoretic stability guarantees throughout learning when utilizing a robust control prior.
	\item We validate experimentally that our algorithm, CORE-RL, exhibits reliably higher performance than the base RL algorithm (and control prior), achieves significant variance reduction in the learning process, and maintains stability throughout learning for stabilization tasks.
\end{itemize}

\section{Related Work}
\label{sec:related}

Significant previous research has examined variance reduction and bias in policy gradient RL. It has been shown that an unbiased estimate of the policy gradient can be obtained from sample trajectories \cite{Williams1992,Sutton1999,Baxter2000}, though these estimates exhibit extremely high variance. This variance can be reduced without introducing bias by subtracting a baseline from the reward function in the policy gradient \cite{Weaver2001,Greensmith2004}. Several works have studied the optimal baseline for variance reduction, often using a critic structure to estimate a value function or advantage function for the baseline \cite{Zhao2012,Silver2014,Schulman2016,Wu2018}. Other works have examined variance reduction in the value function using temporal regularization or regularization directly on the sampled gradient variance \cite{Zhao2015,Thodoroff2018}. However, even with these tools, variance still remains problematically high in reinforcement learning \cite{Islam2017,Henderson2017}. Our work aims to achieve significant further variance reduction \textit{directly on the policy} using control-based \textit{functional} regularization.

Recently, there has been increased interest in functional regularization of deep neural networks, both in reinforcement learning and other domains. Work by \citet{Le2016} has utilized functional regularization to guarantee smoothness of learned functions, and \citet{Benjamin2018} studied properties of functional regularization to limit function distances, though they relied on pointwise sampling from the functions which can lead to high regularizer variance. In terms of utilizing control priors, work by \citet{Johannink2018} adds a control prior during learning, and empirically demonstrates improved performance. Researchers in \citet{Farshidian2014,Nagabandi2017} used model-based priors to produce a good \textit{initialization} for their RL algorithm, but did not use regularization during learning.

Another thread of related work is that of safe RL. Several works on model-based RL have looked at constrained learning such that stability is always guaranteed using Lyapunov-based methods \cite{Perkins2003,Chow2018,Berkenkamp2017}. However, these approaches do not address reward maximization or they overly constrain exploration. On the other hand, work by \citet{Achiam2017} has incorporated constraints (such as stability) into the learning objective, though model-free methods only guarantee approximate constraint satisfaction after a learning period, not during learning \cite{Garcia2015}. Our work proves stability properties throughout learning by taking advantage of the robustness of control-theoretic priors.

\section{Problem Formulation}
\label{sec:problem}

Consider an infinite-horizon discounted Markov decision process (MDP) with deterministic dynamics defined by the tuple $(S, A, f, r, \gamma)$, where $S$ is a set of states, $A$ is a continuous and convex action space, and $f : S \times A \rightarrow S$ describes the system dynamics, which is unknown to the learning agent. The evolution of the system is given by the following dynamical system and its continuous-time analogue,
\begin{equation}
\begin{split}
s_{t+1} & = f(s_t, a_t) = f^{known}(s_t, a_t) + f^{unknown}(s_t, a_t) , \\
\dot{s} &= f_c(s, a) = f_c^{known}(s, a) + f_c^{unknown}(s, a), 
 \end{split}
\label{eq:dynamics_affine}
\end{equation}
where $f^{known}$ captures the known dynamics, $f^{unknown}$ represents the unknowns, $\dot{s}$ denotes the continuous time-derivative of the state $s$, and $f_c(s,a)$ denotes the continuous-time analogue of the discrete time dynamics $f(s_t,a_t)$. A control prior can typically be designed from the known part of the system model, $f^{known}$. 

Consider a stochastic policy $\pi_{\theta}(a | s): S \times A \rightarrow [0,1]$ parameterized by $\theta$. RL aims to find the policy (i.e. parameters, $\theta$) that maximizes the expected accumulated reward $J(\theta)$:
\begin{equation}
J(\theta) = \mathbb{E}_{\tau \sim \pi_{\theta}} [\sum_{t=0}^{\infty} \gamma^t r(s_t, a_t) ] .
\label{eq:RL_cost}
\end{equation}
Here $\tau \sim \pi_{\theta}$ is a trajectory $\tau = \{s_t, a_t, ..., s_{t+n}, a_{t+n} \}$ whose actions and states are sampled from the policy distribution $\pi_{\theta}(a|s)$ and the environmental dynamics (\ref{eq:dynamics_affine}), respectively. The function $r(s,a) : S \times A \rightarrow \mathbb{R}$ is the reward function, and $\gamma \in (0,1)$ is the discount factor. 

This work focuses on policy gradient RL methods, which estimate the gradient of the expected return $J(\theta)$ with respect to the policy based on sampled trajectories. We can estimate the gradient, $\nabla_{\theta} J$, as follows \cite{Sutton1999},
\begin{equation}
\begin{split}
& \nabla_{\theta} J(\theta) = ~ \mathbb{E}_{\tau \sim \pi_{\theta}} \Big[  \nabla_{\theta} \log \pi_{\theta}(\tau) Q^{\pi_{\theta}}(\tau)  \Big] \\
& ~~~~~~~~~~~~~ \approx \sum_{i=1}^N \sum_{t=1}^T [ \nabla_{\theta} \log \pi_{\theta}(s_{i,t}, a_{i,t}) Q^{\pi_{\theta}}(s_{i,t},a_{i,t}) ],  \\
\end{split}
\label{eq:policy_gradient}
\belowdisplayskip 0in
\end{equation}
\noindent
where $Q^{\pi_{\theta}}(s, a)=\mathbb{E}_{\tau \sim \pi_{\theta}} \Big[ \sum_{k=0}^{\infty} \gamma^k r(s_{t+k}, a_{t+k}) |s_t=s, a_t=a \Big]$.
With a good Q-function estimate, the term (\ref{eq:policy_gradient}) is a low-bias estimator of the policy gradient, and utilizes the variance-reduction technique of subtracting a baseline from the reward. However, the resulting policy gradient \textit{still} has very high variance with respect to $\theta$, because the expectation in term (\ref{eq:policy_gradient}) must be estimated using a finite set of sampled trajectories. This high variance in the policy gradient, $\textnormal{var}_{\theta} [\nabla_{\theta} J(\theta_k)]$, translates to high variance in the updated policy, $\textnormal{var}_{\theta} [ \pi_{\theta_{k+1}} ]$, as seen below,
\begin{equation}
\begin{split}
& \theta_{k+1} = \theta_k + \alpha \nabla_{\theta} J(\theta_k), \\
& \pi_{\theta_{k+1}} = \pi_{\theta_k} + \alpha \frac{ d \pi_{\theta_k}}{d \theta} \nabla_{\theta} J(\theta_k) + \mathcal{O}(\Delta \theta^2), \\
& \textnormal{var}_{\theta} [ \pi_{\theta_{k+1}} ] \approx \alpha^2 \frac{d \pi_{\theta_k}}{d \theta} \textnormal{var}_{\theta} [\nabla_{\theta} J(\theta_k)] \frac{d \pi_{\theta_k}}{d \theta} ^T ~~~ \textnormal{for} ~~ \alpha \ll 1, 
\end{split}
\label{eq:policy_variance}
\end{equation}
\noindent
where $\alpha$ is the user-defined learning rate. It is important to note that the variance we are concerned about is \textit{with respect to the parameters} $\theta$, not the noise in the exploration process.

\begin{figure}[!h]
	\centering
	\includegraphics[scale=0.24]{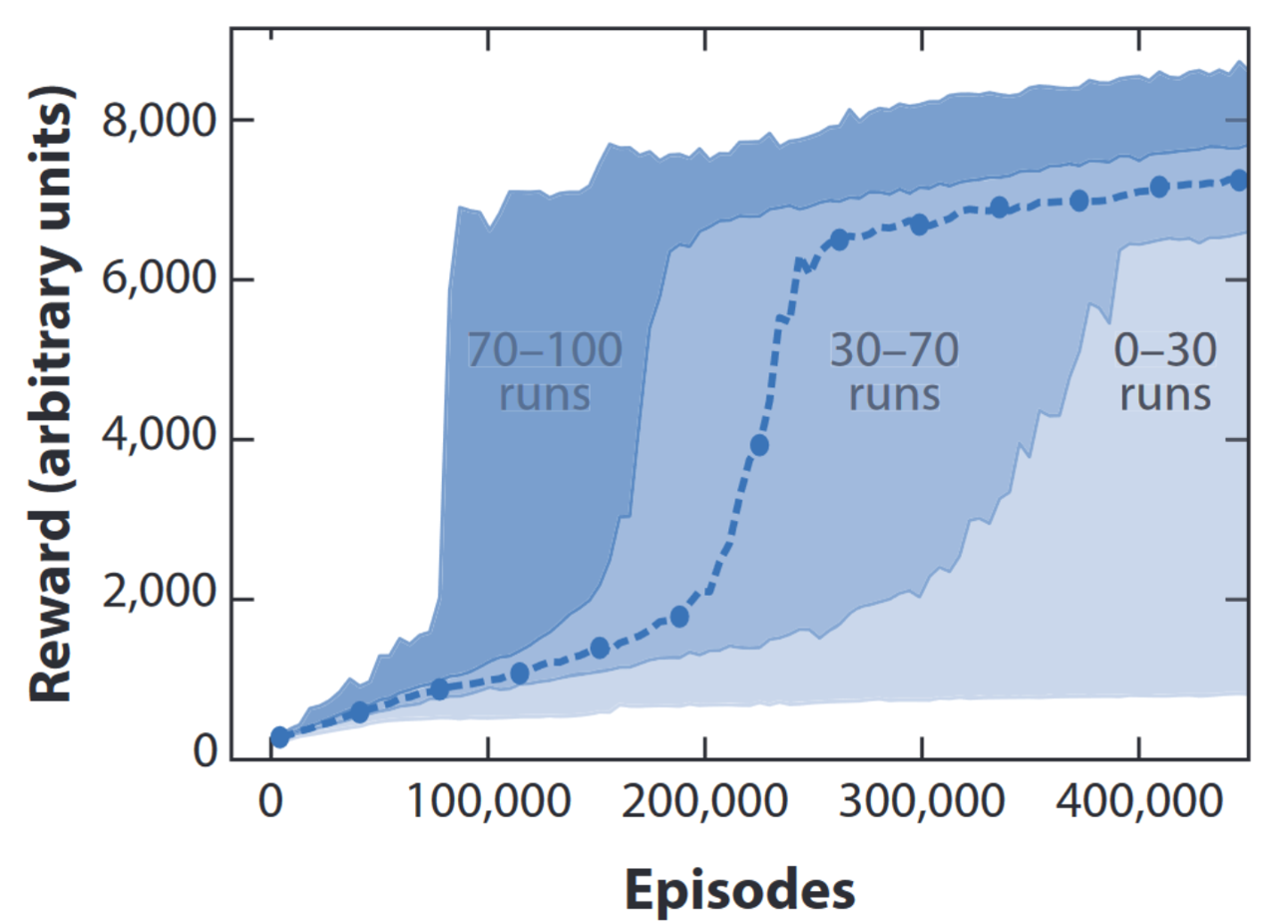}
	\caption{Performance on humanoid walking task from 100 training runs with different initializations. Results from (Recht, 2018).}
	\label{fig:recht_variance}
\end{figure}

To illustrate the variance issue, Fig. \ref{fig:recht_variance} shows the results of 100 separate learning runs using direct policy search on the OpenAI gym task \textit{Humanoid-v1} \cite{Recht2018}. Though high rewards are often achieved, huge variance arises from random initializations and seeds. In this paper, we show that introducing a control prior reduces learning variability, improves learning efficiency, and can provide control-theoretic stability guarantees during learning.

\section{Control Regularization}
\label{sec:frp}

The policy gradient allows us to optimize the objective from sampled trajectories, but it does not utilize  any prior model. However, in many cases we have enough system information to propose at least a crude nominal controller. Therefore, suppose we have a (suboptimal) control prior, $u_{prior}: S \rightarrow A$, and we want to combine our RL policy, $\pi_{\theta_{k}}$, with this control prior at each learning stage, $k$. Before we proceed, let us define $u_{\theta_k}: S \rightarrow A$ to represent the realized controller sampled from the stochastic RL policy $\pi_{\theta_k}(a|s)$ (we will use $u$ to represent deterministic policies and $\pi$ to represent the analogous stochastic ones). We propose to combine the RL policy with the control prior as follows,
\begin{equation}
u_k(s) = \frac{1}{1+\lambda} u_{\theta_k}(s) + \frac{\lambda}{1+\lambda} u_{prior}(s),
\label{eq:policy_mixed}
\end{equation}
where we assume a continuous, convex action space. Note that $u_k(s)$ is the realized controller sampled from stochastic policy $\pi_k$, whose distribution over actions has been shifted by $u_{prior}$ such that $\pi_k \Big( \frac{1}{1+\lambda} a + \frac{\lambda}{1+\lambda} u_{prior} ~ \Big| ~ s \Big) = \pi_{\theta_k}(a|s)$. We refer to $u_k$ as the \textit{mixed policy}, and $u_{\theta_k}$ as the RL policy.

Utilizing the mixed policy (\ref{eq:policy_mixed}) is equivalent to placing a functional regularizer $u_{prior}$ on the RL policy, $u_{\theta_k}$, with regularizer weight $\lambda$. Let $\pi_{\theta_k}(a|s)$ be Gaussian distributed: $\pi_{\theta_k} = \mathcal{N}(\overline{u}_{\theta_k}, \Sigma)$, so that $\Sigma$ describes the exploration noise. Then we obtain the following,
\begin{equation}
\begin{split}
& \overline{u}_k(s) = \frac{1}{1+\lambda} \overline{u}_{\theta_k}(s) +  \frac{\lambda}{1+\lambda} u_{prior}(s),
\end{split}
\label{eq:policy_mix}
\end{equation}
where the control prior, $u_{prior}$ can be interpreted as a Gaussian prior on the mixed control policy (see Appendix A). Let us define the norm $ \| u_1 - u_2 \|_{\Sigma} = (u_1 - u_2)^T \Sigma^{-1} (u_1 - u_2) $.

\begin{lemma}
		The policy $\overline{u}_k(s)$ in Equation (\ref{eq:policy_mix}) is the solution to the following regularized optimization problem,
		\begin{equation}
		\begin{split}
		& \overline{u}_k(s) = \argmin_{u} ~~ \Big\| u(s) -  \overline{u}_{\theta_k} \Big\|_{\Sigma} ~ \\
		& ~~~~~~~~~~~~~~~~~~ + \lambda || u(s) - u_{prior}(s)||_{\Sigma}, ~~~ \forall s \in S, \\
		\end{split}
		\label{eq:opt_reg_lemma}
		\end{equation}
		\noindent
		which can be equivalently expressed as the constrained optimization problem,
		\begin{equation}
		\begin{split}
		& \overline{u}_k(s) = \argmin_{u} ~~ \Big\| u(s) -  \overline{u}_{\theta_k} \Big\|_{\Sigma} \\
		& ~~~~~ \textnormal{s.t.} ~~~~  ||u (s) - u_{prior} (s) ||_{\Sigma} \leq \tilde{\mu}(\lambda) ~~~~ \forall s \in S , \\
		\end{split}
		\label{eq:opt_constraint}
		\end{equation}
		\noindent
		where $\tilde{\mu}$ constrains the policy search. Assuming convergence of the RL algorithm, $\overline{u}_k(s)$ converges to the solution, 
		\begin{equation}
		\begin{split}
		& \overline{u}_{k}(s) = \argmin_{u} ~~ \Big\| u(s) -  \argmax_{\bar{u}_{\theta}} \mathbb{E}_{\tau \sim \bar{u}} \Big[ r(s,a) \Big] \Big\|_{\Sigma} ~ \\
		& ~~~~~~ + \lambda || u(s) - u_{prior}(s)||_{\Sigma}, ~~~ \forall s \in S ~~ \textnormal{as } ~  k \rightarrow \infty \\
		\end{split}
		\label{eq:opt_reg1_lemma}
		\end{equation} 
\end{lemma}
This lemma is proved in Appendix A. The equivalence between (\ref{eq:policy_mix}) and (\ref{eq:opt_reg_lemma}) illustrates that the control prior acts as a functional regularization (recall that $\overline{u}_{\theta_k}$ solves the reward maximization problem appearing in (\ref{eq:opt_reg1_lemma}) ). The policy mixing  (\ref{eq:policy_mix}) can \textit{also} be interpreted as constraining policy search near the control prior, as shown by (\ref{eq:opt_constraint}). More weight on the control prior (higher $\lambda$) constrains the policy search more heavily. In certain settings, the problem can be solved in the constrained optimization formulation \cite{Le2019}.

\subsection{CORE-RL Algorithm}

Our learning algorithm is described in Algorithm \ref{alg:mix_1}. At the high level, the process can be described as:
\begin{itemize}[topsep=0pt, leftmargin=*]
	\itemsep-0.2em
	\item First compute the control prior based on prior knowledge (Line 1). See Section 5 for details on controller synthesis.
	\item For a given policy iteration, compute the regularization weight, $\lambda$, using the strategy described in Section 4.3 (Lines 7-9). The algorithm can also use a fixed regularization weight, $\lambda$ (Lines 10-11).
	\item Deploy the mixed policy (\ref{eq:policy_mixed}) on the system, and record the resulting states/action/rewards (Lines 13-15). 
	\item At the end of each policy iteration, update the policy based on the recorded state/action/rewards (Lines 16-18).
\end{itemize}

\begin{algorithm}[tb]
	\caption{Control Regularized RL (CORE-RL)}\label{alg:mix_1}
	\begin{algorithmic}[1]
		\STATE Compute the control prior, $u_{prior}$ using the known model $f^{known}(s,a)$ (or other prior knowledge)
		\STATE Initialize RL policy $\pi_{\theta_0}$
		\STATE Initialize array $\mathcal{D}$ for storing rollout data
		\STATE Set $k=1$ (representing $k^{th}$ policy iteration)
		\WHILE {$k < ~ $Episodes}
		\STATE Evaluate policy $\pi_{\theta_{k-1}}$ at each timestep
		\IF{Using Adaptive Mixing Strategy}
		\STATE At each timestep, compute regularization weight $\lambda$ 
		\STATE{~ for the control prior using the TD-error from (\ref{eq:lambda_td}).}
		\ELSE
		\STATE Set constant regularization weight $\lambda$
		\ENDIF
		\STATE Deploy mixed policy $\pi_{k-1}$ from (\ref{eq:policy_mixed}) to obtain 
		\STATE{~~~~~ rollout of state-action-reward for T timesteps.}
		\STATE{Store resulting data ($s_t, a_t, r_t, s_{t+1}$) in array $\mathcal{D}$.}
		\STATE Using data in $\mathcal{D}$, update policy using any policy 
		\STATE{~~~~~ gradient-based RL algorithm (e.g. DDPG, PPO)}
		\STATE{~~~~~to obtain $\theta_k$.}
		\STATE $k = k + 1$
		\ENDWHILE
		\STATE \textbf{return} Policy $\pi_{\theta_k}, ~ u_{prior} ~~~~~~~~~~~~~~~~ \rhd$ Overall controller
	\end{algorithmic}
\end{algorithm}

\subsection{Bias-Variance Tradeoff}

Theorem \ref{theorem:bias_variance} formally states that mixing the policy gradient-based controller, $\pi_{\theta_k}$, with the control prior, $u_{prior}$, decreases learning variability. However, the mixing may introduce bias into the learned policy that depends on the (a) regularization $\lambda$, and (b) sub-optimality of the control prior. Bias is defined in (\ref{eq:policy_bias_theorem}) and refers to the difference between the mixed policy and the (potentially locally) optimal RL policy \textit{at convergence}.
\begin{theorem}
	Consider the mixed policy (\ref{eq:policy_mixed}) where $\pi_{\theta_k}$ is a policy gradient-based RL policy, and denote the (potentially local) optimal policy to be $\pi_{opt}$. The variance (\ref{eq:policy_variance}) of the mixed policy arising from the policy gradient is reduced by a factor $(\frac{1}{1+\lambda})^2$ when compared to the RL policy with no control prior. 
	
	However, the mixed policy may introduce bias proportional to the sub-optimality of the control prior. If we let $D_{sub} = D_{TV}(\pi_{opt}, \pi_{prior})$, then the policy bias (i.e. $D_{TV} ( \pi_{k}, \pi_{opt} )$) is bounded as follows,
	\begin{equation}
	\begin{split}
	D_{TV} & ( \pi_{k}, \pi_{opt} ) \geq D_{sub} - \frac{1}{1 + \lambda} D_{TV}  ( \pi_{\theta_k}, \pi_{prior} ) \\
	D_{TV} & ( \pi_{k}, \pi_{opt} ) \leq \frac{\lambda}{1 + \lambda} D_{sub} ~~~~~~ \textnormal{as  } k \rightarrow \infty \\
	\end{split}
	\label{eq:policy_bias_theorem}
	\end{equation}
	\noindent
	where $D_{TV}(\cdot, \cdot)$ represents the total variation distance between two probability measures (i.e. policies). Thus, if $D_{sub}$ and $\lambda$ are large, this will introduce policy bias.
	
	\label{theorem:bias_variance}
\end{theorem}

The proof can be found in Appendix B. Recall that $\pi_{prior}$ is the stochastic analogue to the deterministic control prior $u_{prior}$, such that $\pi_{prior}(a|s) = \mathds{1}(a = u_{prior}(s))$ where $\mathds{1}$ is the indicator function. Note that the bias/variance results apply to the \textit{policy} -- not the accumulated reward. 

\textbf{Intuition:} Using Figure 2, we provide some intuition for the control regularization discussed above. Note the following:

\begin{enumerate}[topsep=0pt, leftmargin=*, label=\arabic*)]
	\itemsep-0.17em
	\item The explorable region of the state space is denoted by the set $\mathcal{S}_{st}$, which grows as $\lambda$ decreases and vice versa. This illustrates the constrained policy search interpretation of regularization in the state space.
	\item The difference between the control prior trajectory and optimal trajectory (i.e. $D_{sub}$) \textit{may} bias the final policy depending on the explorable region $\mathcal{S}_{st}$ (i.e. $\lambda$). Fig 2. shows this difference, and its implications, in state space.
	\item If the optimal trajectory is within the explorable region, then we can learn the corresponding optimal policy -- otherwise the policy will remain suboptimal. 
\end{enumerate}
Points 1 and 3 will be formally addressed in Section 5.

\begin{figure}[!h]
	\centering
	\includegraphics[scale=0.34]{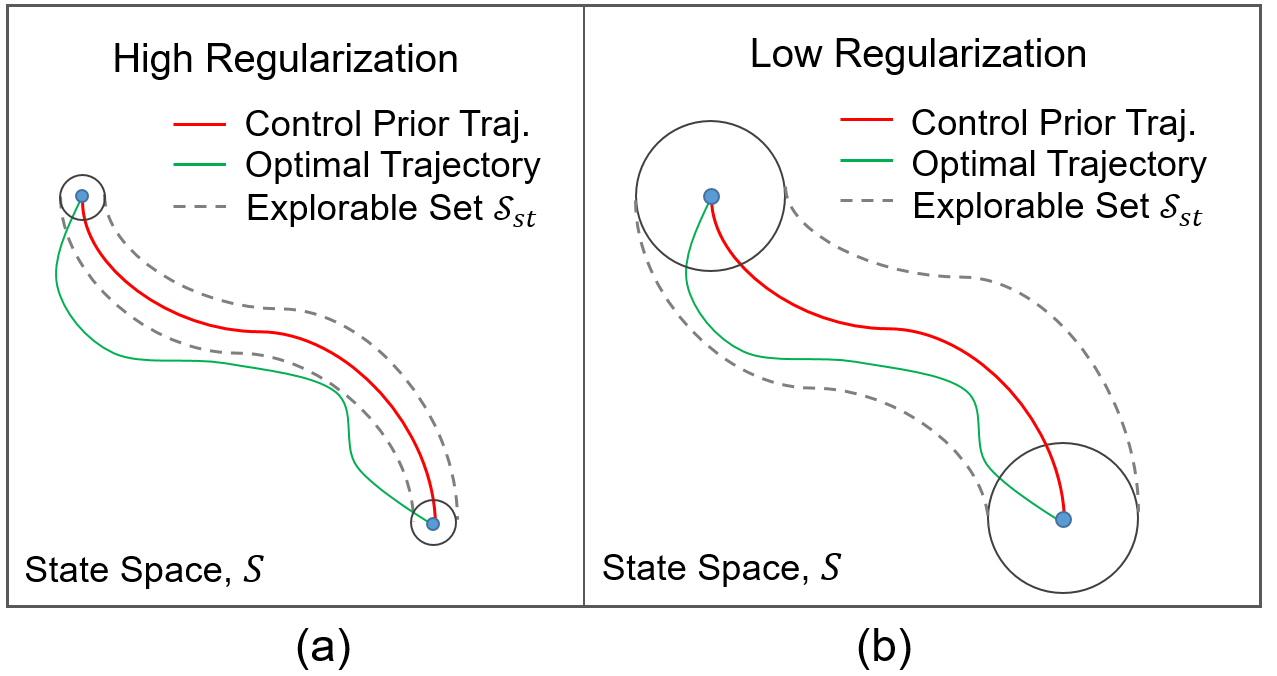}
	\caption{Illustration of optimal trajectory vs. control-theoretic trajectory with the explorable set $\mathcal{S}_{st}$. (a) With high regularization, set $\mathcal{S}_{st}$ is small so we cannot learn the optimal trajectory. (b) With lower regularization, set $\mathcal{S}_{st}$ is larger so we \textit{can} learn the optimal trajectory. However, this also enlarges the policy search space.}
	\label{fig:stability_goal}
\end{figure}

\subsection{Computing the mixing parameter $\lambda$}

A remaining challenge is automatically tuning $\lambda$, especially as we acquire more training data. While setting a fixed $\lambda$ can perform well, intuitively, $\lambda$ should be large when the RL controller is highly uncertain, and it should decrease as we become more confident in our learned controller.

Consider the multiple model adaptive control (MMAC) framework, where a set of controllers (each based on a different underlying model) are generated. A meta-controller computes the overall controller by selecting the weighting for different candidate controllers, based on how close the underlying system model for each candidate controller is to the ``true'' model \cite{Kuipers2010}. Inspired by this approach, we should weight the RL controller proportional to our confidence in its model. Our confidence should be state-dependent (i.e. low confidence in areas of the state space where little data has been collected). However, since the RL controller does not utilize a dynamical system model, we propose measuring confidence in the RL controller via the magnitude of the \textit{temporal difference} (TD) error,
\begin{equation}
| \delta^{\pi}(s_t) | = | r_{t+1} + \gamma Q^{\pi}(s_{t+1}, a_{t+1}) - Q^{\pi}(s_t, a_t) |, 
\label{eq:td_error}
\end{equation}
\noindent
where $a_t \sim \pi(a | s_t) , ~ a_{t+1} \sim \pi(a|s_{t+1})$. This TD error measures how poorly the RL algorithm predicts the value of subsequent actions from a given state. A high TD-error implies that the estimate of the action-value function at a given state is poor, so we should rely more heavily on the control prior (a high $\lambda$ value). In order to scale the TD-error to a value in the interval $\lambda \in [0,\lambda_{max}]$, we take the negative exponential of the TD-error, computed at run-time,
\begin{equation}
\lambda(s_t) = \lambda_{max} \Big( 1 - e^{- C | \delta(s_{t-1}) | } \Big). \\
\label{eq:lambda_td}
\end{equation}
\noindent
The parameters $C$ and $\lambda_{max}$ are tuning parameters of the adaptive weighting strategy. Note that Equation (\ref{eq:lambda_td}) uses $\delta({s_{t-1}})$ rather than $\delta({s_{t}})$, because computing $\delta({s_{t}})$ requires measurement of state $s_{t+1}$. Thus we rely on the reasonable assumption that $\delta(s_t) \approx \delta(s_{t-1})$, since $s_t$ should be very close to $s_{t-1}$ in practice. 

Equation (\ref{eq:lambda_td}) yields a low value of $\lambda$ if the RL action-value function predictions are accurate. This measure is chosen because the (explicit) underlying model of the RL controller is the value function (rather than a dynamical system model). Our experiments show that this adaptive scheme based on the TD error allows better tuning of the variance and performance of the policy.

\section{Control Theoretic Stability Guarantees}
\label{sec:barrier_guided}

In many controls applications, it is crucial to ensure dynamic stability, not just high rewards, during learning. When a (crude) dynamical system model is available, we can utilize classic controller synthesis tools (i.e. LQR, PID, etc.) to obtain a stable control prior in a region of the state space. In this section, we utilize a well-established tool from robust control theory ($\mathcal{H}^{\infty}$ control), to analyze system stability under the mixed policy (\ref{eq:policy_mixed}), and prove stability guarantees throughout learning when using a robust control prior.

Our work is built on the idea that the control prior should \textit{maximize robustness to disturbances and model uncertainty}, so that we can treat the RL control, $u_{\theta_k}$, as a performance-maximizing ``disturbance'' to the control prior, $u_{prior}$. The mixed policy then takes advantage of the stability properties of the robust control prior, \textit{and} the performance optimization properties of the RL algorithm. To obtain a robust control prior, we utilize concepts from $\mathcal{H}^{\infty}$ control \cite{Doyle1996}. 

Consider the nonlinear dynamical system (\ref{eq:dynamics_affine}), and let us linearize the \textit{known} part of the model $f_c^{known}(s,a)$ around a desired equilibrium point to obtain the following,
\begin{equation}
\begin{split}
\dot{s} & = A s + B_1 w + B_2 a \\
z & = C_1 s + D_{11} w + D_{12} a
\end{split}
\label{eq:dynamics_linear}
\end{equation}
where $w \in \mathbb{R}^{m_1}$ is the disturbance vector, and $z \in \mathbb{R}^{p_1}$ is the controlled output. Note that we analyze the continuous-time dynamics rather than discrete-time, since all mechanical systems have continuous time dynamics that can be discovered through analysis of the system Lagrangian. However, similar analysis can be done for discrete-time dynamics. We make the following standard assumption -- conditions for its satisfaction can be found in \cite{Doyle1989},

\begin{assumption} A $\mathcal{H}^{\infty}$ controller exists for linear system ($\ref{eq:dynamics_linear}$) that stabilizes the system in a region of the state space.
\end{assumption}
Stability here means that system trajectories are bounded around the origin/setpoint. We can then synthesize an $H^{\infty}$ controller, $u^{\mathcal{H}^{\infty}}(s) = -K s$, using established techniques described in \cite{Doyle1989}. The resulting controller is robust with \textit{worst-case disturbances} attenuated by a factor $\zeta_k$ before entering the output, where $\zeta_k < 1$ is a parameter returned by the synthesis algorithm. See Appendix F for further details on $\mathcal{H}^{\infty}$ control and its robustness properties.

Having synthesized a robust $\mathcal{H}^{\infty}$ controller for the \textit{linear system model} (\ref{eq:dynamics_linear}), we are interested in how those robustness properties (e.g. disturbance attenuation by $\zeta_k$) influence the nonlinear system (\ref{eq:dynamics_affine}) controlled by the mixed policy (\ref{eq:policy_mixed}). We rewrite the system dynamics (\ref{eq:dynamics_affine}) in terms of the linearization (\ref{eq:dynamics_linear}) plus a disturbance term as follows,
\begin{equation}
\dot{s} = f_c(s,a) = A s + B_2 a + d(s,a),
\label{eq:dynamics_linear_uncertain}
\end{equation}
where $d(s,a)$ gathers together all dynamic uncertainties and nonlinearities. To keep this small, we could use feedback linearization based on the nominal nonlinear model (\ref{eq:dynamics_affine}).

We now analyze stability of the nonlinear system (\ref{eq:dynamics_linear_uncertain}) under the mixed policy (\ref{eq:policy_mixed}) using Lyapunov analysis \cite{Khalil2000}. Consider the Lyapunov function $V(s) = s^T P s$, where $P$ is obtained when synthesizing the $\mathcal{H}^{\infty}$ controller (see Appendix F). If we can define a closed region, $\mathcal{S}_{st}$, around the origin such that $\dot{V}(s) < 0$ outside that region, then by standard Lyapunov analysis, $\mathcal{S}_{st}$ is forward invariant and asymptotically stable (note $\dot{V}(s)$ is the time-derivative of the Lyapunov function). Since the $\mathcal{H}^{\infty}$ control law satisfies an Algebraic Riccati Equation, we obtain the following relation,

\begin{lemma}
	For any state $s$, satisfaction of the condition,
	\begin{equation*}
	\begin{split}
	2 s^T P \Big( d(s,a) + \frac{1}{1+\lambda} & B_2 u_e \Big)  < \\
	& s^T (C_1^T C_1 + \frac{1}{\zeta_k^2}PB_1 B_1^T P) s ,
	\end{split}
	\label{eq:H_Lyapunov_mix_robust}
	\end{equation*}
	implies that $\dot{V}(s) < 0$.
\end{lemma}

This lemma is proved in Appendix C. Note that $u_{e} =u_{\theta} - u^{\mathcal{H}^{\infty}}$ denotes the difference between the RL controller and control prior, and $(A, B_1, B_2, C_1)$ come from (\ref{eq:dynamics_linear}). Let us bound the RL control output such that $\| u_{e} \|_2 \leq C_{\pi}$, and define the set $\mathcal{C} = \{(s,u) \in (S,A) ~ \Big| ~ \| u_{e} \|_2 \leq C_{\pi} , ~ H^{\infty} ~ \textnormal{control is stabilizing} \}$. We also bound the ``disturbance'' $\| d(s,a) \|_2 \leq C_D$, for all $s \in \mathcal{C}$, and define the minimum singular value $\sigma_{m} (\zeta_k) = \sigma_{min}(C_1^T C_1 + \frac{1}{\zeta^2_k} P B_1 B_1^T P) $, which reflects the robustness of the control prior (i.e. larger $\sigma_{m}$ imply greater robustness). Then using Lemma 2 and Lyapunov analysis tools, we can derive a conservative set that is guaranteed asymptotically stable and forward invariant under the mixed policy, as described in the following theorem (proof in Appendix D).

\begin{theorem}
	Assume a stabilizing $H^{\infty}$ control prior within the set $\mathcal{C}$ for the dynamical system (\ref{eq:dynamics_linear_uncertain}). Then asymptotic stability and forward invariance of the set $\mathcal{S}_{st} \subseteq \mathcal{C}$
	\begin{equation}
	\begin{split}
	&  \mathcal{S}_{st}: \{s \in \mathbb{R}^n: \| s \|_2 \leq \frac{1}{\sigma_{m} (\zeta_k)} \Big( 2 \| P \|_2 C_D  \\ 
	& ~~~~~~~~~~~~~~~~~~~~~~ + \frac{2}{1+\lambda} \| P B_2 \|_2 C_{\pi} \Big) ~ , ~ s \in \mathcal{C} \}.
	\end{split}
	\label{eq:stable_set}
	\end{equation}
	is guaranteed under the mixed policy (\ref{eq:policy_mixed}) for all $s \in \mathcal{C}$. The set $\mathcal{S}_{st}$ contracts as we (a) increase robustness of the control prior (increase $\sigma_m(\zeta_k)$), (b) decrease our dynamic uncertainty/nonlinearity $C_D$, or (c) increase weighting $\lambda$ on the control prior.
\end{theorem}

Put simply, Theorem 2 says that all states in $\mathcal{C}$ will converge to (and remain within) set $\mathcal{S}_{st}$ under the mixed policy (\ref{eq:policy_mixed}). Therefore, the stability guarantee is stronger if $\mathcal{S}_{st}$ has smaller cardinality. The set $\mathcal{S}_{st}$ is drawn pictorally in Fig. 2, and essentially dictates the explorable region. Note that $\mathcal{S}_{st}$ is \textit{not} the region of attraction.

Theorem 2 highlights the tradeoff between robustness parameter, $\zeta_k$, of the control prior, the nonlinear uncertainty in the dynamics $C_D$, and the utilization of the learned controller, $\lambda$. If we have a more robust control prior (higher $\sigma_m(\zeta_k)$) or better knowledge of the dynamics (smaller $C_D$), we can heavily weight the learned controller (lower $\lambda$) during the learning process while still guaranteeing stability.

While shrinking the set $\mathcal{S}_{st}$ and achieving asymptotic stability along a trajectory or equilibrium point may seem desirable, Fig. \ref{fig:stability_goal} illustrates why this is not necessarily the case in an RL context. The optimal trajectory for a task typically deviates from the nominal trajectory (i.e. the control theoretic-trajectory), as shown in Fig. \ref{fig:stability_goal} -- the set $\mathcal{S}_{st}$ illustrates the explorable region under regularization. Fig. \ref{fig:stability_goal}(a) shows that we do not want strict stability of the nominal trajectory, and instead would like \textit{limited} flexibility (a sufficiently large $\mathcal{S}_{st}$) to explore. By increasing the weighting on the learned policy $\pi_{\theta_k}$ (decreasing $\lambda$), we expand the set $\mathcal{S}_{st}$ and allow for greater exploration around the nominal trajectory (at the cost of stability) as seen in Fig. \ref{fig:stability_goal}(b).

\section{Empirical Results}
\label{sec:results}

We apply the CORE-RL Algorithm to three problems: (1) cartpole stabilization, (2) car-following control with experimental data, and (3) racecar driving with the TORCS simulator. We show results using DDPG or PPO or TRPO \cite{Lillicrap2015, Schulman2017, Schulman2015} as the policy gradient RL algorithm (PPO + TRPO results moved to Appendix G), though any similar RL algorithm could be used. All code can be found at \url{https://github.com/rcheng805/CORE-RL}.

Note that our results focus on reward rather than bias. Bias (as defined in Section 4.2) assumes convergence to a (locally) optimal policy, and does not include many factors influencing performance (e.g. slow learning, failure to converge, etc.). In practice, Deep-RL algorithms often do not converge (or take very long to do so). Therefore, reward better demonstrates the influence of control regularization on performance, which is of greater practical interest.



\subsection{CartPole Problem}

We apply the CORE-RL algorithm to control of the cartpole from the OpenAI gym environment (\textit{CartPole-v1}). We modified the CartPole environment so that it takes a \textit{continuous} input, rather than discrete input, and we utilize a reward function that encourages the cartpole to maintain its $x-$position while keeping the pole upright. Further details on the environment and reward function are in Appendix E. To obtain a control prior, we assume a crude model (i.e. linearization of the nonlinear dynamics with $\approx60\%$ error in the mass and length values), and from this we synthesize an $\mathcal{H}^{\infty}$ controller. Using this control prior, we run Algorithm 1 with several different regularization weights, $\lambda$. For each $\lambda$, we run CORE-RL 6 times with different random seeds.

Figure \ref{fig:results}a plots reward improvement over the control prior, which shows that the regularized controllers perform much better than the baseline DDPG algorithm (in terms of variance, reward, and learning speed). We also see that intermediate values of $\lambda$ (i.e. $\lambda \approx 4$) result in the best learning, demonstrating the importance of policy regularization. 

Figure \ref{fig:results}b better illustrates the performance-variance tradeoff. For small $\lambda$, we see high variance \textit{and} poor performance. With intermediate $\lambda$, we see higher performance and lower variance. As we further increase $\lambda$, variance continues to decrease, but the performance also decreases since policy exploration is heavily constrained. The adaptive mixing strategy performs very well, exhibiting low variance through learning, and converging on a high-performance policy.

\begin{figure}[!h]
	\centering
	\includegraphics[scale=0.37]{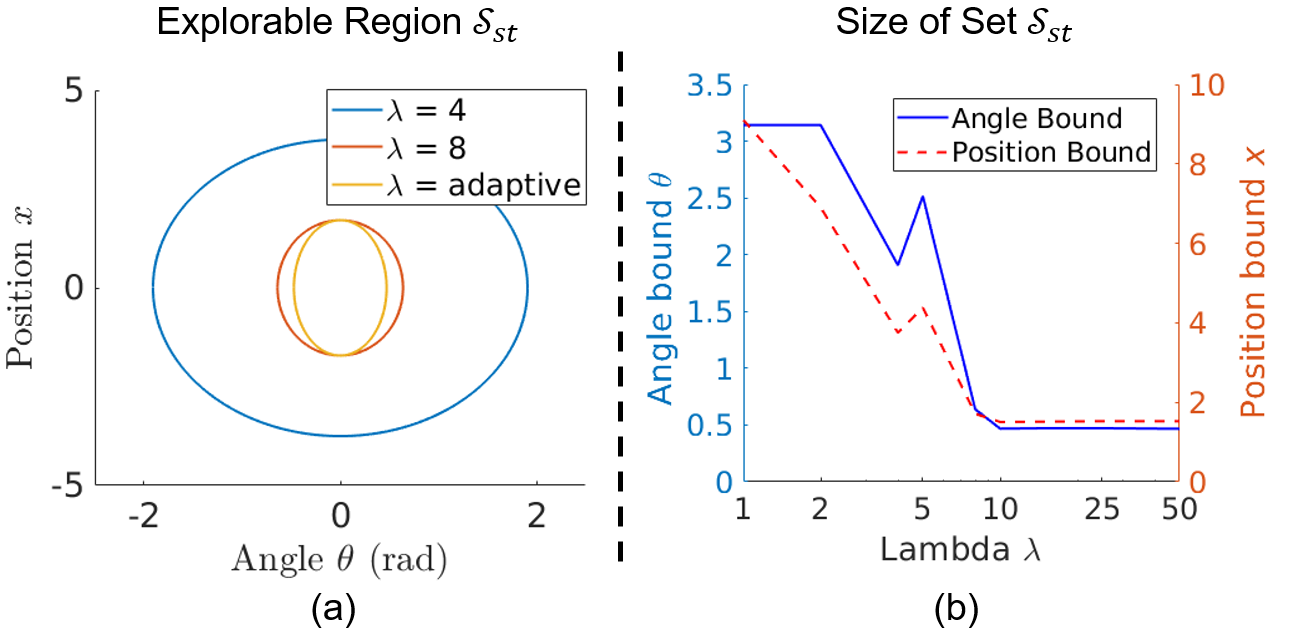}
	\caption{Stability region for CartPole under mixed policy. (a) Illustration of the stability region for different regularization, $\lambda$. For each $\lambda$ shown, the trajectory goes to and remains within the corresponding stability set throughout training. (b) Size of the stability region in terms of the angle $\theta$, and position $x$. As $\lambda$ increases, we are guaranteed to remain closer to the equilibrium point during learning.}
	\label{fig:stability_cartpole}
\end{figure}

While Lemma 1 proved that the mixed controller (\ref{eq:policy_mix}) has the same \textit{optimal} solution as optimization problem (\ref{eq:opt_reg_lemma}), when we ran experiments directly using the loss in (\ref{eq:opt_reg_lemma}), we found that performance (i.e. reward) was worse than CORE-RL and still suffered high variance. In addition, learning with pre-training on the control prior likewise exhibited high variance and had worse performance than CORE-RL.

\begin{figure*}[!h]
	\centering
	\includegraphics[scale=0.61]{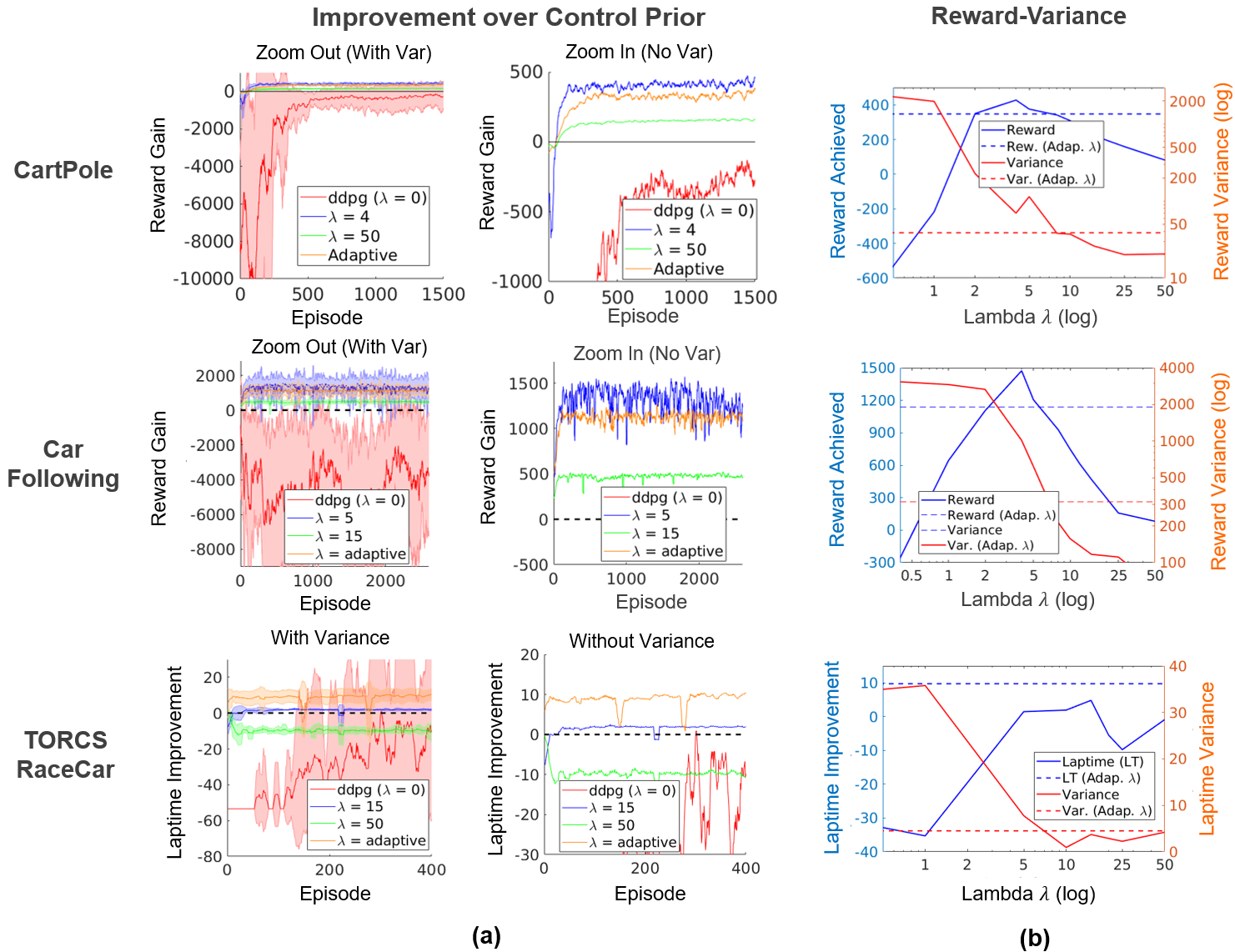}
	\caption{Learning results for CartPole, Car-Following, and TORCS RaceCar Problems using DDPG. (a) Reward improvement over control prior with different set values for $\lambda$ or an adaptive $\lambda$. The right plot is a zoomed-in version of the left plot without variance bars for clarity. Values above the dashed black line signify improvements over the control prior. (b) Performance and variance in the reward as a function of the regularization $\lambda$, across different runs of the algorithm using random initializations/seeds. Dashed lines show the performance (i.e. reward) and variance using the adaptive weighting strategy. Variance is measured for all episodes across all runs. Adaptive $\lambda$ and intermediate values of $\lambda$ exhibit best learning. Again, performance is baselined to the control prior, so any performance value above 0 denotes improvement over the control prior.}
	\label{fig:results}
\end{figure*}

Importantly, according to Theorem 2, the system should maintain stability (i.e. remain within an invariant set around our desired equilibrium point) \textit{throughout the learning process}, and the stable region shrinks as we increase $\lambda$. Our simulations exhibit exactly this property as seen in Figure \ref{fig:stability_cartpole}, which shows the \textit{maximum} deviation from the equilibrium point across all episodes. The system converges to a stability region throughout learning, and this region contracts as we increase $\lambda$. Therefore, regularization not only improves learning performance and decreases variance, but can capture stability guarantees from a robust control prior.

\subsection{Experimental Car-Following}

We next examine experimental data from a chain of 5 cars following each other on an 8-mile segment of a single-lane public road. We obtain position (via GPS), velocity, and acceleration data from each of the cars, and we control the acceleration/deceleration of the $4^{th}$ car in the chain. The goal is to learn an optimal controller for this car that maximizes fuel efficiency while avoiding collisions. The experimental setup and data collection process are described in \cite{Ge2018}. For the control prior, we utilize a bang-bang controller that (inefficiently) tries to maintain a large distance from the car in front and behind the controlled car. The reward function penalizes fuel consumption and collisions (or near-collisions). Specifics of the control prior, reward function, and experiments are in Appendix E.

For our experiments, we split the data into 10 second ``episodes'', shuffle the episodes, and run CORE-RL six times with different random seeds (for several different $\lambda$). 

Figure \ref{fig:results}a shows again that the regularized controllers perform much better than the baseline DDPG algorithm for the car-following problem, and demonstrates that regularization leads to performance improvements over the control prior and gains in learning efficiency. Figure \ref{fig:results}b reinforces that intermediate values of $\lambda$ (i.e. $\lambda \approx 5$) exhibit optimal performance. Low values of $\lambda$ exhibit significant deterioration of performance, because the car must learn (with few samples) in a much larger policy search space; the RL algorithm does not have enough data to converge on an optimal policy. High values of $\lambda$ also exhibit lower performance because they heavily constrain learning. Intermediate $\lambda$ allow for the best learning using the limited number of experiments.

Using an adaptive strategy for setting $\lambda$ (or alternatively tuning to an optimal $\lambda$), we obtain high-performance policies that improve upon both the control prior and RL baseline controller. The variance is also low, so that the learning process \textit{reliably} learns a good controller.

\subsection{Driving in \torcs}

Finally we run CORE-RL to generate controllers for cars in \emph{The Open Racing Car Simulator} (\torcs) \cite{TORCS}. The simulator provides readings from $29$ sensors, which describe the environment state. The sensors provide information like car speed, distance from track center, wheel spin, etc. The controller decides values for the acceleration, steering and braking actions taken by the car.

To obtain a control prior for this environment, we use a simple PID-like linearized controller for each action, similar to the one described in \cite{PIRL}. These types of controllers are known to have sub-optimal performance, while still being able to drive the car around a lap. We perform all our experiments on the CG-Speedway track in \torcs. For each $\lambda$, we run the algorithm $5$ times with different initializations and random seeds.

For \textsc{Torcs}, we plot \textit{laptime improvement over the control prior} so that values above zero denote improved performance over the prior. The laps are timed out at $150$s, and the objective is to minimize lap-time by completing a lap as fast as possible. Due to the sparsity of the lap-time signal, we use a pseudo-reward function during training that provides a heuristic estimate of the agent's performance at each time step during the simulation (details in Appendix E).

Once more, Figure \ref{fig:results}a shows that regularized controllers perform better on average than the baseline DDPG algorithm, and that we improve upon the control prior with proper regularization. Figure \ref{fig:results}b shows that intermediate values of $\lambda$ exhibit good performance, but using the adaptive strategy for setting $\lambda$ in the TORCS setting gives us the highest-performance policy that significantly beats both the control prior and DDPG baseline. Also, the variance with the adaptive strategy is significantly lower than for the DDPG baseline, which again shows that the learning process \textit{reliably} learns a good controller.

Note that we have only shown results for DDPG. Results for PPO and TRPO are similar for CartPole and Car-following (different for TORCS), and can be found in Appendix G.

\section{Conclusion}
\label{sec:conclusion}

A significant criticism of RL is that random seeds can produce vastly different learning behaviors, limiting application of RL to real systems. This paper shows, through theoretical results and experimental validation, that our method of control regularization substantially alleviates this problem, enabling significant variance reduction and performance improvements in RL. This regularization can be interpreted as constraining the explored action space during learning. 

Our method also allows us to capture dynamic stability properties of a robust control prior to guarantee stability during learning, and has the added benefit that it can easily incorporate different RL algorithms (e.g. PPO, DDPG, etc.). The main limitation of our approach is that it relies on a reasonable control prior, and it remains to be analyzed how bad of a control prior can be used while still aiding learning.

\section*{Acknowledgements}
This work was funded in part by Raytheon under the Learning to Fly program, and by DARPA under the Physics-Infused AI Program.


\bibliographystyle{icml2019}
\bibliography{references}

\begin{thebibliography}{38}
\providecommand{\natexlab}[1]{#1}
\providecommand{\url}[1]{\texttt{#1}}
\expandafter\ifx\csname urlstyle\endcsname\relax
  \providecommand{\doi}[1]{doi: #1}\else
  \providecommand{\doi}{doi: \begingroup \urlstyle{rm}\Url}\fi

\bibitem[Achiam et~al.(2017)Achiam, Held, Tamar, and Abbeel]{Achiam2017}
Achiam, J., Held, D., Tamar, A., and Abbeel, P.
\newblock {Constrained Policy Optimization}.
\newblock In \emph{International Conference on Machine Learning (ICML)}, 2017.

\bibitem[Arulkumaran et~al.(2017)Arulkumaran, Deisenroth, Brundage, and
  Bharath]{Arulkumaran2017}
Arulkumaran, K., Deisenroth, M.~P., Brundage, M., and Bharath, A.~A.
\newblock {Deep reinforcement learning: A brief survey}.
\newblock \emph{IEEE Signal Processing Magazine}, 2017.

\bibitem[Baxter \& Bartlett(2000)Baxter and Bartlett]{Baxter2000}
Baxter, J. and Bartlett, P.
\newblock {Reinforcement learning in POMDP's via direct gradient ascent}.
\newblock \emph{International Conference on Machine Learning}, 2000.

\bibitem[Benjamin et~al.(2018)Benjamin, Rolnick, and Kording]{Benjamin2018}
Benjamin, A.~S., Rolnick, D., and Kording, K.
\newblock {Measuring and Regularizing Networks in Function Space}.
\newblock \emph{arXiv:1805.08289}, 2018.

\bibitem[Berkenkamp et~al.(2017)Berkenkamp, Turchetta, Schoellig, and
  Krause]{Berkenkamp2017}
Berkenkamp, F., Turchetta, M., Schoellig, A.~P., and Krause, A.
\newblock {Safe Model-based Reinforcement Learning with Stability Guarantees}.
\newblock In \emph{Neural Information Processing Systems (NeurIPS)}, 2017.

\bibitem[Chow et~al.(2018)Chow, Nachum, Duenez-Guzman, and
  Ghavamzadeh]{Chow2018}
Chow, Y., Nachum, O., Duenez-Guzman, E., and Ghavamzadeh, M.
\newblock {A Lyapunov-based Approach to Safe Reinforcement Learning}.
\newblock In \emph{Advances in Neural Information Processing Systems
  (NeurIPS)}, 2018.

\bibitem[Doyle(1996)]{Doyle1996}
Doyle, J.
\newblock {Robust and Optimal Control}.
\newblock In \emph{Conference on Decision and Control}, 1996.

\bibitem[Doyle et~al.(1989)Doyle, Glover, Khargonekar, and Francis]{Doyle1989}
Doyle, J., Glover, K., Khargonekar, P., and Francis, B.
\newblock {State-space solutions to standard H/sub 2/ and H/sub infinity /
  control problems}.
\newblock \emph{IEEE Transactions on Automatic ControlTransactions on Automatic
  Control}, 1989.
\newblock ISSN 00189286.
\newblock \doi{10.1109/9.29425}.

\bibitem[Duan et~al.(2016)Duan, Chen, Schulman, and Abbeel]{Duan2016}
Duan, Y., Chen, X., Schulman, J., and Abbeel, P.
\newblock {Benchmarking Deep Reinforcement Learning for Continuous Control}.
\newblock In \emph{International Conference on Machine Learning (ICML)}, 2016.

\bibitem[Farshidian et~al.(2014)Farshidian, Neunert, and
  Buchli]{Farshidian2014}
Farshidian, F., Neunert, M., and Buchli, J.
\newblock {Learning of closed-loop motion control}.
\newblock In \emph{IEEE International Conference on Intelligent Robots and
  Systems}, 2014.

\bibitem[Garc{\'{i}}a \& Fern{\'{a}}ndez(2015)Garc{\'{i}}a and
  Fern{\'{a}}ndez]{Garcia2015}
Garc{\'{i}}a, J. and Fern{\'{a}}ndez, F.
\newblock {A Comprehensive Survey on Safe Reinforcement Learning}.
\newblock \emph{JMLR}, 2015.

\bibitem[Ge et~al.(2018)Ge, Avedisov, He, Qin, Sadeghpour, and Orosz]{Ge2018}
Ge, J.~I., Avedisov, S.~S., He, C.~R., Qin, W.~B., Sadeghpour, M., and Orosz,
  G.
\newblock {Experimental validation of connected automated vehicle design among
  human-driven vehicles}.
\newblock \emph{Transportation Research Part C: Emerging Technologies}, 2018.

\bibitem[Ghosh et~al.(2018)Ghosh, Singh, Rajeswaran, Kumar, and
  Levine]{Ghosh2018}
Ghosh, D., Singh, A., Rajeswaran, A., Kumar, V., and Levine, S.
\newblock Divide-and-conquer reinforcement learning.
\newblock In \emph{Neural Information Processing Systems (NeurIPS)}, volume
  abs/1711.09874, 2018.

\bibitem[Greensmith et~al.(2004)Greensmith, Bartlett, and
  Baxter]{Greensmith2004}
Greensmith, E., Bartlett, P., and Baxter, J.
\newblock {Variance reduction techniques for gradient estimates in
  reinforcement learning}.
\newblock \emph{JMLR}, 2004.

\bibitem[Henderson et~al.(2018)Henderson, Islam, Bachman, Pineau, Precup, and
  Meger]{Henderson2017}
Henderson, P., Islam, R., Bachman, P., Pineau, J., Precup, D., and Meger, D.
\newblock {Deep Reinforcement Learning that Matters}.
\newblock In \emph{AAAI Conference on Artificial Intelligence}, 2018.

\bibitem[Islam et~al.(2017)Islam, Henderson, Gomrokchi, and Precup]{Islam2017}
Islam, R., Henderson, P., Gomrokchi, M., and Precup, D.
\newblock {Reproducibility of Benchmarked Deep Reinforcement Learning of Tasks
  for Continuous Control}.
\newblock In \emph{Reproducibility in Machine Learning Workshop}, 2017.

\bibitem[{Johannink} et~al.(2018){Johannink}, {Bahl}, {Nair}, {Luo}, {Kumar},
  {Loskyll}, {Aparicio Ojea}, {Solowjow}, and {Levine}]{Johannink2018}
{Johannink}, T., {Bahl}, S., {Nair}, A., {Luo}, J., {Kumar}, A., {Loskyll}, M.,
  {Aparicio Ojea}, J., {Solowjow}, E., and {Levine}, S.
\newblock {Residual Reinforcement Learning for Robot Control}.
\newblock \emph{arXiv e-prints}, art. arXiv:1812.03201, Dec 2018.

\bibitem[Khalil(2000)]{Khalil2000}
Khalil, H.~K.
\newblock \emph{{Nonlinear Systems (Third Edition)}}.
\newblock Prentice Hall, 2000.

\bibitem[Kuipers \& Ioannou(2010)Kuipers and Ioannou]{Kuipers2010}
Kuipers, M. and Ioannou, P.
\newblock {Multiple model adaptive control with mixing}.
\newblock \emph{IEEE Transactions on Automatic Control}, 2010.

\bibitem[Le et~al.(2016)Le, Kang, Yue, and Carr]{Le2016}
Le, H., Kang, A., Yue, Y., and Carr, P.
\newblock {Smooth Imitation Learning for Online Sequence Prediction}.
\newblock In \emph{International Conference on Machine Learning (ICML)}, 2016.

\bibitem[Le et~al.(2019)Le, Voloshin, and Yue]{Le2019}
Le, H.~M., Voloshin, C., and Yue, Y.
\newblock Batch policy learning under constraints.
\newblock In \emph{International Conference on Machine Learning}, 2019.

\bibitem[Lillicrap et~al.(2016)Lillicrap, Hunt, Pritzel, Heess, Erez, Tassa,
  Silver, and Wierstra]{Lillicrap2015}
Lillicrap, T.~P., Hunt, J.~J., Pritzel, A., Heess, N., Erez, T., Tassa, Y.,
  Silver, D., and Wierstra, D.
\newblock {Continuous control with deep reinforcement learning}.
\newblock In \emph{International Conference on Learning Representations}, 2016.

\bibitem[{Nagabandi} et~al.(2017){Nagabandi}, {Kahn}, {Fearing}, and
  {Levine}]{Nagabandi2017}
{Nagabandi}, A., {Kahn}, G., {Fearing}, R.~S., and {Levine}, S.
\newblock {Neural Network Dynamics for Model-Based Deep Reinforcement Learning
  with Model-Free Fine-Tuning}.
\newblock \emph{arXiv e-prints}, art. arXiv:1708.02596, Aug 2017.

\bibitem[Perkins \& Barto(2003)Perkins and Barto]{Perkins2003}
Perkins, T.~J. and Barto, A.~G.
\newblock {Lyapunov design for safe reinforcement learning}.
\newblock \emph{Journal of Machine Learning Research}, 2003.

\bibitem[Recht(2019)]{Recht2018}
Recht, B.
\newblock {A Tour of Reinforcement Learning: The View from Continuous Control}.
\newblock \emph{Annual Review of Control, Robotics, and Autonomous Systems},
  2\penalty0 (1):\penalty0 253--279, 2019.

\bibitem[Schulman et~al.(2015)Schulman, Levine, Moritz, Jordan, and
  Abbeel]{Schulman2015}
Schulman, J., Levine, S., Moritz, P., Jordan, M., and Abbeel, P.
\newblock {Trust Region Policy Optimization}.
\newblock In \emph{International Conference on Machine Learning (ICML)}, 2015.

\bibitem[Schulman et~al.(2016)Schulman, Moritz, Levine, Jordan, and
  Abbeel]{Schulman2016}
Schulman, J., Moritz, P., Levine, S., Jordan, M., and Abbeel, P.
\newblock {High-Dimensional Continuous Control Using Generalized Advantage
  Estimation}.
\newblock \emph{International Conference on Learning Representations}, 2016.

\bibitem[{Schulman} et~al.(2017){Schulman}, {Wolski}, {Dhariwal}, {Radford},
  and {Klimov}]{Schulman2017}
{Schulman}, J., {Wolski}, F., {Dhariwal}, P., {Radford}, A., and {Klimov}, O.
\newblock {Proximal Policy Optimization Algorithms}.
\newblock \emph{arXiv e-prints}, art. arXiv:1707.06347, Jul 2017.

\bibitem[Silver et~al.(2014)Silver, Lever, Heess, Degris, Wierstra, and
  Riedmiller]{Silver2014}
Silver, D., Lever, G., Heess, N., Degris, T., Wierstra, D., and Riedmiller, M.
\newblock {Deterministic Policy Gradient Algorithms}.
\newblock \emph{Proceedings of the 31st International Conference on Machine
  Learning (ICML-14)}, 2014.

\bibitem[Sutton et~al.(1999)Sutton, McAllester, Singh, and Mansour]{Sutton1999}
Sutton, R., McAllester, D., Singh, S.~P., and Mansour, Y.
\newblock {Policy Gradient Methods for Reinforcement Learning with Function
  Approximation}.
\newblock \emph{Advances in Neural Information Processing Systems}, 1999.

\bibitem[Thodoroff et~al.(2018)Thodoroff, Durand, Pineau, and
  Precup]{Thodoroff2018}
Thodoroff, P., Durand, A., Pineau, J., and Precup, D.
\newblock {Temporal Regularization for Markov Decision Process}.
\newblock In \emph{Advances in Neural Information Processing Systems}, 2018.

\bibitem[Verma et~al.(2018)Verma, Murali, Singh, Kohli, and Chaudhuri]{PIRL}
Verma, A., Murali, V., Singh, R., Kohli, P., and Chaudhuri, S.
\newblock Programmatically interpretable reinforcement learning.
\newblock In \emph{International Conference on Machine Learning (ICML)}, 2018.

\bibitem[Weaver \& Tao(2001)Weaver and Tao]{Weaver2001}
Weaver, L. and Tao, N.
\newblock {The Optimal Reward Baseline for Gradient-Based Reinforcement
  Learning}.
\newblock In \emph{Uncertainty in Artificial Intelligence (UAI)}, 2001.

\bibitem[Williams(1992)]{Williams1992}
Williams, R.~J.
\newblock {Simple statistical gradient-following algorithms for connectionist
  reinforcement learning}.
\newblock \emph{Machine Learning}, 1992.

\bibitem[Wu et~al.(2018)Wu, Rajeswaran, Duan, Kumar, Bayen, Kakade, Mordatch,
  and Abbeel]{Wu2018}
Wu, C., Rajeswaran, A., Duan, Y., Kumar, V., Bayen, A.~M., Kakade, S.,
  Mordatch, I., and Abbeel, P.
\newblock {Variance Reduction for Policy Gradient with Action-Dependent
  Factorized Baselines}.
\newblock In \emph{International Conference on Learning Representations}, 2018.

\bibitem[Wymann et~al.(2014)Wymann, Espi{\'e}, Guionneau, Dimitrakakis, Coulom,
  and Sumner]{TORCS}
Wymann, B., Espi{\'e}, E., Guionneau, C., Dimitrakakis, C., Coulom, R., and
  Sumner, A.
\newblock {TORCS}, {T}he {O}pen {R}acing {C}ar {S}imulator.
\newblock http://www.torcs.org, 2014.

\bibitem[Zhao et~al.(2012)Zhao, Hachiya, Niu, and Sugiyama]{Zhao2012}
Zhao, T., Hachiya, H., Niu, G., and Sugiyama, M.
\newblock {Analysis and improvement of policy gradient estimation}.
\newblock \emph{Neural Networks}, 2012.

\bibitem[Zhao et~al.(2015)Zhao, Niu, Xie, Yang, and Sugiyama]{Zhao2015}
Zhao, T., Niu, G., Xie, N., Yang, J., and Sugiyama, M.
\newblock {Regularized Policy Gradients : Direct Variance Reduction in Policy
  Gradient Estimation}.
\newblock \emph{Proceedings of the Asian Conference on Machine Learning}, 2015.

\end{thebibliography}

\newpage
\twocolumn[
\icmltitle{Appendix: Control Regularization for Reduced Variance Reinforcement Learning}
]

\appendix

\setcounter{theorem}{0}
\setcounter{lemma}{0}

\section{Proof of Lemma 1}

\begin{lemma}
	The policy $\overline{u}_k(s)$ in Equation (\ref{eq:policy_mix}) is the solution to the following regularized optimization problem,
	\begin{equation}
	\begin{split}
	& \overline{u}_k(s) = \argmin_{u} ~~ \Big\| u(s) -  \overline{u}_{\theta_k} \Big\|^2 ~ \\
	& ~~~~~~~~~~~~~~~~~~ + \lambda || u(s) - u_{prior}(s)||^2, ~~~ \forall s \in S, \\
	\end{split}
	\label{eq:opt_reg_lemma_appendix}
	\end{equation}
	\noindent
	which can be equivalently expressed as the constrained optimization problem:
	\begin{equation}
	\begin{split}
	& \overline{u}_k(s) = \argmin_{u} ~~ \Big\| u(s) -  \overline{u}_{\theta_k} \Big\|^2 \\
	& ~~~~~ \textnormal{s.t.} ~~~~  ||u (s) - u_{prior} (s) ||^2 \leq \tilde{\mu}(\lambda) ~~~~ \forall s \in S , \\
	\end{split}
	\label{eq:opt_constraint_lemma}
	\end{equation}
	\noindent
	where $\tilde{\mu}$ constrains the policy search. Assuming convergence of the RL algorithm, $\overline{u}_k(s)$ converges to the solution, 
	\begin{equation}
	\begin{split}
	& \overline{u}_{k}(s) = \argmin_{u} ~~ \Big\| u(s) -  \argmax_{u_{\theta}} \mathbb{E}_{\tau \sim u} \Big[ r(s,a) \Big] \Big\|^2 ~ \\
	& ~~~~~~~~ + \lambda || u(s) - u_{prior}(s)||^2, ~~~ \forall s \in S ~~~ \textnormal{as } ~  k \rightarrow \infty \\
	\end{split}
	\label{eq:opt_reg1_lemma_appendix}
	\end{equation}
\end{lemma}

\begin{proof}
	~
	
	\textbf{Equivalence between (\ref{eq:policy_mix}) and (\ref{eq:opt_reg_lemma_appendix}) : }
	Let $\pi_{\theta_k}(a|s)$ be a Gaussian distributed policy with mean $\overline{u}_{\theta_k}(s)$: ~ $\pi_{\theta_k}(a | s) \sim \mathcal{N}(\overline{u}_{\theta_k}(s), \Sigma)$. Thus, $\Sigma$ describes exploration noise. From the mixed policy definition (\ref{eq:policy_mix}), we can obtain the following Gaussian distribution describing the mixed policy:
	\begin{equation}
	\begin{split}
	& \pi_k(a|s) = \mathcal{N} (\frac{1}{1+\lambda} \overline{u}_{\theta_k} + \frac{1}{1+\lambda} u_{prior}, \Sigma ) 
	\\
	& ~~~ = \frac{1}{c_N} \mathcal{N} \Big( \overline{u}_{\theta_k}(s) , (1+\lambda) \Sigma \Big) \cdot \mathcal{N} \Big( u_{prior}(s) , \frac{1+\lambda}{\lambda} \Sigma \Big),
	\end{split}
	\label{eq:gaussian_product}
	\end{equation}
	where the second equality follows based on the properties of products of Gaussians. Let us define $ \| u_1 - u_2 \|_{\Sigma} = (u_1 - u_2)^T \Sigma^{-1} (u_1 - u_2) $, and let $| \Sigma |$ be the determinant of $|\Sigma|$. Then, distribution (\ref{eq:gaussian_product}) can be rewritten as the product,
	\begin{equation}
	\begin{split}
	& \mathbb{P}(X(s)) = - c_1 \exp( - \frac{1}{2(1+\lambda)}\| X(s) - \overline{u}_{\theta_k}(s) \|_{\Sigma} ) ~ \times \\
	& ~~~~~~~~~~~~~~~~~ - c_1 \lambda^{\frac{k}{2}} \exp( -\frac{\lambda}{2(1+\lambda)}\| X(s) - u_{prior}(s) \|_{\Sigma} ) \\
	& ~~~~~~~ c_1 = \frac{1}{c_N \sqrt{(2 \pi)^k (1 + \lambda)^k | \Sigma |}}
	\end{split}
	\label{eq:bayes_gaussian}
	\end{equation}
	
	\noindent
	where $X(s)$ is a random variable with $\mathbb{P}(X(s))$ representing the probability of taking action $X$ from state $s$ under policy (\ref{eq:policy_mix}). Further simplifying this PDF, we obtain:
	\begin{equation}
	\begin{split}
	& \mathbb{P}(X(s)) = c_2 \exp \Big( - \| X(s) - \overline{u}_{\theta_k}(s) \|_{\Sigma} \\
	& ~~~~~~~~~~~~~~~~~~~~~~~~~~~~~~~~~~~~ - \lambda \| X(s) - u_{prior}(s) \|_{\Sigma} \Big) \\
	& ~~~~~~~~~~~ c_2 = \frac{\lambda^{\frac{k}{2}}}{c_N (2 \pi)^k (1 + \lambda)^k | \Sigma |}
	\end{split}
	\label{eq:bayes_gaussian1}
	\end{equation}
		
	Since the probability $\mathbb{P}(X(s))$ is maximized when the argument of the exponential in Equation (\ref{eq:bayes_gaussian1}) is minimized, then the maximum probability policy can be expressed as the solution to the following regularized optimization problem,
	\begin{equation}
	\begin{split}
	& \overline{u}_k(s) = \argmin_{u}(s) ~~ \| u(s) - \overline{u}_{\theta_k}(s) \|_{\Sigma} ~ + \\
	& ~~~~~~~~~~~~~~~~~~ \lambda \| u(s) - u_{prior}(s)\|_{\Sigma}, ~~~~~ \forall s \in S . \\
	\end{split}
	\label{eq:opt_reg_1}
	\end{equation}
	
	\noindent
	Therefore the mixed policy $\overline{u}_{k}(s)$ from Equation (\ref{eq:policy_mix}) is the solution to Problem (\ref{eq:opt_reg_lemma_appendix}) .
	
	~
	
	\textbf{Convergence of (\ref{eq:opt_reg_lemma_appendix}) to  (\ref{eq:opt_reg1_lemma_appendix}): } Note that $\overline{u}_{\theta_k}$ and $\pi_{\theta_k}$ are parameterized by the same $\theta_k$ and represent the iterative solution to the optimization problem $\argmax_{\theta} \mathbb{E}_{\tau \sim u_{k}} \Big[ r(\tau) \Big]$ at the latest policy iteration. Thus, assuming convergence of the RL algorithm, we can rewrite problem (\ref{eq:opt_reg_1}) as follows,	
	\begin{equation}
	\begin{split}
	& \overline{u}_{k} = \argmin_{u} ~~ \Big\| u(s) -  \argmax_{u_{\theta_k}} \mathbb{E}_{\tau \sim u_{k}} \Big[ r(s,a) \Big] \Big\|^2 \\
	& ~~~~~~~~~~~~~~~~~~ + \lambda || u(s) - u_{prior}(s)||^2, ~~~ \forall s \in S.\\
	\end{split}
	\label{eq:opt_reg_2}
	\end{equation}
	
	~
	
	\textbf{Equivalence between (\ref{eq:opt_reg_lemma_appendix}) and (\ref{eq:opt_constraint_lemma}) : }
	Finally, we want to show that the solutions for regularized problem (\ref{eq:opt_reg_lemma_appendix}) and the constrained optimization problem (\ref{eq:opt_constraint_lemma}) are equivalent. 
	
	First, note that Problem (\ref{eq:opt_reg_lemma_appendix}) is the dual to Problem (\ref{eq:opt_constraint_lemma}), where $\lambda$ is the dual variable. Clearly problem (\ref{eq:opt_reg_lemma_appendix}) is convex in $u$. Furthermore, Slater's condition holds, since there is always a feasible point (e.g. trivially $u(s) = u_{prior}(s)$). Therefore strong duality holds. This means that $\exists \lambda \geq 0$ such that the solution to Problem (\ref{eq:opt_constraint_lemma}) must also be optimal for Problem (\ref{eq:opt_reg_lemma_appendix}).
	
	To show the other direction, fix $\lambda > 0$ and define $R(u) = \| u(s) - \overline{u}_{\theta_k}(s) \|^2$ and $C(u) = || u(s) - u_{prior}(s)||^2$ for all $s \in S$. Let us denote $u^*$ as the optimal solution for Problem (\ref{eq:opt_reg_lemma_appendix}) with $C(u^*) = \tau > \tilde{\mu}$ (note we can choose $\tilde{\mu}$). However supposed $u^*$ is \textit{not} optimal for Problem (\ref{eq:opt_constraint_lemma}). Then there exists $\tilde{u}$ such that $R(u^*) < R(\tilde{u} )$ and $C(\tilde{u}) \leq \tilde{\mu}$. Denote the difference in the two rewards by $ R(\tilde{u}) - R(u^*) = R_{diff}$. Thus the following relations hold,
	\begin{equation}
	\begin{split}
	R(\tilde{u}) + \lambda C(\tilde{u}) < R(u^*) + \lambda C(u^*) + R_{diff} + \lambda \big[ \tilde{\mu} - \tau \big].
	\end{split}
	\label{eq:opt_constraint_lemma_appendix}
	\end{equation}
	This leads to the conditional statement,
	\begin{equation}
	\begin{split}
	& ~ R_{diff} + \lambda \big[ \tilde{\mu} - \tau \big] \geq 0 \\
	&  ~~~~~~~~~~ \Rightarrow ~~~ R(\tilde{u}) + \lambda C(\tilde{u}) < R(u^*) + \lambda C(u^*).
	\end{split}
	\label{eq:opt_constraint_lemma_appendix1}
	\end{equation}
	For fixed $\lambda$, there always exists $\tilde{\mu} > 0$ such that the condition $R_{diff} + \lambda \big[ \tilde{\mu} - \tau \big] \geq 0$ holds. However, this leads to a contradiction, since we assumed that $u^*$ is optimal for Problem (\ref{eq:opt_reg_lemma_appendix}). We can conclude then that $\exists \tilde{\mu}$ such that the solution to Problem (\ref{eq:opt_reg_lemma_appendix}) must be optimal for Problem (\ref{eq:opt_constraint_lemma}). Therefore, Problems (\ref{eq:opt_reg_lemma_appendix}) and (\ref{eq:opt_constraint_lemma}) have equivalent solutions.
	
\end{proof}

\section{Proof of Theorem 1}

\begin{theorem}
	Consider the mixed policy (\ref{eq:policy_mixed}) where $\pi_{\theta_k}$ is an RL controller learned through policy gradients, and denote the (potentially local) optimal policy to be $\pi_{opt}$. The variance (\ref{eq:policy_variance}) of the mixed policy arising from the policy gradient is reduced by a factor $(\frac{1}{1+\lambda})^2$ when compared to the RL policy with no control prior. 
	
	However, the mixed policy may introduce bias proportional to the sub-optimality of the control prior. More formally, if we let $D_{sub} = D_{TV}(\pi_{opt}, \pi_{prior})$, then the policy bias (i.e. $D_{TV} ( \pi_{k}, \pi_{opt} )$) is bounded as follows:
	\begin{equation}
	\begin{split}
	D_{TV} & ( \pi_{k}, \pi_{opt} ) \geq D_{sub} - \frac{1}{1 + \lambda} D_{TV}  ( \pi_{\theta_k}, \pi_{prior} ) \\
	D_{TV} & ( \pi_{k}, \pi_{opt} ) \leq \frac{\lambda}{1 + \lambda} D_{sub} ~~~~~~ \textnormal{as  } k \rightarrow \infty \\
	\end{split}
	\label{eq:policy_bias_theorem_appendix}
	\end{equation}
	\noindent
	where $D_{TV}(\cdot, \cdot)$ represents the total variation distance between two probability measures (i.e. policies). Thus, if $D_{sub}$ and $\lambda$ are large, this will introduce policy bias.
	
	\label{theorem:bias_variance_appendix}
\end{theorem}

\begin{proof}
	Let us define the stochastic action (i.e. random variable) $\mathcal{A}^{act}_{k+1} \sim \pi_{\theta_{k+1}}(a|s)$. Then recall from Equation (\ref{eq:policy_variance}) that assuming a fixed, Gaussian distributed policy, $\pi_{\theta_k}(a|s)$,
	\begin{equation}
	\textnormal{var}_{\theta}[\mathcal{A}^{act}_{k+1} | s]  \approx \alpha^2 \frac{d \pi_{\theta_k}}{d \theta} \textnormal{var}_{\theta} [\nabla_{\theta} J(\theta_k)] \frac{d \pi_{\theta_k}}{d \theta} ^T .
	\label{eq:policy_var_proof}
	\end{equation}
	Based on the mixed policy definition (\ref{eq:policy_mixed}), we obtain the following relation between the variance of $\pi_k$ and $\pi_{\theta_k}$ (the mixed policy and RL policy, respectively),
	\begin{equation}
	\begin{split}
	\textnormal{var}_{\theta} [ \pi_{k+1} ] &= \textnormal{var}_{\theta} \Big[ \frac{1}{1+\lambda} \mathcal{A}^{act}_{k+1} + \frac{\lambda}{1+\lambda} u_{prior} | s \Big] \\ 
	&= \frac{1}{(1+\lambda)^2} \textnormal{var}_{\theta} [ \mathcal{A}^{act}_{k+1} | s ] \\
	&= \frac{\alpha^2}{(1+\lambda)^2} \frac{d \pi_{\theta_k}}{d \theta} \textnormal{var}_{\theta} [\nabla_{\theta} J(\theta_k)] \frac{d \pi_{\theta_k}}{d \theta} ^T.  \\
	\end{split}
	\label{eq:policy_var_mix}
	\end{equation}
	Compared to the variance (\ref{eq:policy_variance}), we achieve a variance reduction when utilizing the same learning rate $\alpha$. Taking the same policy gradient from (\ref{eq:policy_variance}), $\textnormal{var} [\nabla_{\theta} J(\theta_k)]$, then the variance is reduced by a factor of $(\frac{1}{1+\lambda})^2$ by introducing policy mixing.
	
	Lower variance comes at a price -- potential introduction of bias into policy. Let us define the policy bias as $D_{TV} ( \pi_{k}, \pi_{opt} )$, and let us denote $D_{sub} = D_{TV}(\pi_{opt}, \pi_{prior})$. Since total variational distance, $D_{TV}$ is a metric, we can use the triangle inequality to obtain:
	\begin{equation}
	\begin{split}
	&  D_{TV} ( \pi_{k} , \pi_{opt} ) \geq D_{TV} ( \pi_{prior} , \pi_{opt} ) - D_{TV} ( \pi_{prior} , \pi_{k} ).
	\end{split}
	\label{eq:policy_bias1_appendix}
	\end{equation}
	We can further break down the term $D_{TV} ( \pi_{prior} , \pi_{k})$: 
	\begin{equation}
	\begin{split}
	D_{TV} & (\pi_{prior}, \pi_{k})  \\
	&= \sup_{(s,a) \in  S \textnormal{x} A} \Big| \pi_{prior} - \frac{1}{1+\lambda} \pi_{\theta_{k}} - \frac{\lambda}{1+\lambda} \pi_{prior} \Big| \\
	&= \frac{1}{1+\lambda} \sup_{(s,a) \in  S \textnormal{x} A} | \pi_{\theta_k} - \pi_{prior} | \\
	& = \frac{1}{1+\lambda} D_{TV} (\pi_{\theta_k} , \pi_{prior}).
	\end{split}
	\label{eq:policy_bias2_appendix}
	\end{equation}
	This holds for all $k \in \mathbb{N}$. From (\ref{eq:policy_bias1_appendix}) and (\ref{eq:policy_bias2_appendix}), we can obtain the lower bound in (\ref{eq:policy_bias_theorem_appendix}),
	\begin{equation*}
	\begin{split}
	D_{TV} & ( \pi_{k}, \pi_{opt} ) \geq D_{sub} - \frac{1}{1 + \lambda} D_{TV}  ( \pi_{\theta_k}, \pi_{prior} ) \\
	\end{split}
	\label{eq:policy_bias3_appendix}
	\end{equation*}
	
	To obtain the upper bound, let the policy gradient algorithm with \textit{no} control prior achieve asymptotic convergence to the (locally) optimal policy $\pi_{opt}$ (as proven for certain classes of function approximators in \cite{Sutton1999}). Denote this policy as $\pi^{(p)}_{\theta_k}$, such that $\pi^{(p)}_{\theta_k} \rightarrow \pi_{opt}$ as $k \rightarrow \infty$. In this case, we can derive the total variation distance between the mixed policy (\ref{eq:policy_mixed}) and the optimal policy as follows,
	\begin{equation}
	\begin{split}
	D_{TV} & ( \pi_{opt} , \pi^{(p)}_{k} )  \\
	&= \sup_{(s,a) \in  S \textnormal{x} A} | \pi_{opt} - \frac{1}{1+\lambda} \pi^{(p)}_{\theta_{k}} - \frac{\lambda}{1+\lambda} \pi_{prior} | \\
	&= \frac{\lambda}{1+\lambda} \sup_{(s,a) \in  S \textnormal{x} A} | \pi_{opt} - \pi_{prior} | ~~~~ \textnormal{as } k \rightarrow \infty \\
	& = \frac{\lambda}{1+\lambda} D_{TV} (\pi_{opt} , \pi_{prior}) ~~~~ \textnormal{as } k \rightarrow \infty \\
	& = \frac{\lambda}{1+\lambda} D_{sub} ~~~~ \textnormal{as } k \rightarrow \infty .
	\end{split}
	\label{eq:policy_bias_lower}
	\end{equation}
	Note that this represents an \textit{upper bound} on the bias, since it assumes that $\pi^{(p)}_{\theta_k}$ is uninfluenced by $\pi_{prior}$ during learning. It shows that $\pi^{(p)}_{\theta_k}$ is a feasible policy, but not necessarily optimal when accounting for regularization with $\pi_{prior}$. Therefore, we can obtain the upper bound:
	\begin{equation}
	\begin{split}
	D_{TV} ( \pi_{opt} , \pi_{k} ) & \leq D_{TV} ( \pi_{opt} , \pi^{(p)}_{k} ) \\
	& = \frac{\lambda}{1+\lambda} D_{sub} ~~~ \textnormal{as } k \rightarrow \infty .   
	\end{split}
	\label{eq:policy_bias_upper}
	\end{equation}

\end{proof}

\section{Proof of Lemma 2}
	\begin{lemma}
	For any state $s$, satisfaction of the condition,
	\begin{equation*}
	\begin{split}
	2 s^T P \Big( d(s,a) + \frac{1}{1+\lambda} & B_2 u_e \Big)  < \\
	& s^T (C_1^T C_1 + \frac{1}{\gamma_k^2}PB_1 B_1^T P) s ,
	\end{split}
	\label{eq:H_Lyapunov_mix_robust_appendix}
	\end{equation*}
	implies that $\dot{V}(s) < 0$.
    \end{lemma}

	\begin{proof}
		Recall that we are analyzing the Lyapunov function $V(s) = s^T P s$, where P is taken from the Algebraic Riccati Equation (\ref{eq:riccatti_appendix}). Let us take the time derivative of the Lyapunov function as follows:
        \begin{equation}
        \begin{split}
        \dot{V}(s) &= \frac{dV}{ds} \dot{s} = 2 s^T P \Big( A s + B_2 a + d(s,a) \Big) \\
        &= s^T (-C_1^T C_1 - \frac{1}{\gamma_k^2}PB_1 B_1^T P) s + 2 s^T P d(s,a) + \\
        & ~~~~~~~~~~~ \frac{2}{1+\lambda} s^T P B_2 (u_{\theta_k} - u_{prior}) \\
        &= s^T (-C_1^T C_1 - \frac{1}{\gamma_k^2}PB_1 B_1^T P) s + 2 s^T P \Big( d(s,a) + \\
        & ~~~~~~~~~~~ \frac{1}{1+\lambda} B_2 u_e \Big) .
        \end{split}
        \label{eq:H_Lyapunov_appendix_appendix}
        \end{equation}
    The second equality comes from the Algebraic Riccati Equation (\ref{eq:riccatti_appendix}), which the dynamics satisfy by design of the $\mathcal{H}^{\infty}$ controller. From here, it follows directly that if, 
    \begin{equation*}
	\begin{split}
	2 s^T P \Big( d(s,a) + \frac{1}{1+\lambda} & B_2 u_e \Big)  < \\
	& s^T (C_1^T C_1 + \frac{1}{\gamma_k^2}PB_1 B_1^T P) s ,
	\end{split}
	\end{equation*}
	then $\dot{V}(s) < 0$.
	\end{proof}
		
\section{Proof of Theorem 2}

\begin{theorem}
	Assume a stabilizing $H^{\infty}$ control prior within the set $\mathcal{C}$ for our dynamical system (\ref{eq:dynamics_linear_uncertain}). Then asymptotic stability and forward invariance of the set $\mathcal{S}_{st} \subseteq \mathcal{C}$
	\begin{equation}
	\begin{split}
	&  \mathcal{S}_{st}: \{s \in \mathbb{R}^n: \| s \|_2 \leq \frac{1}{\sigma_{m} (\gamma_k)} \Big( 2 \| P \|_2 C_D  \\ 
	& ~~~~~~~~~~~~~~~~~~~~~~ + \frac{2}{1+\lambda} \| P B_2 \|_2 C_{\pi} \Big) ~ , ~ s \in \mathcal{C} \}.
	\end{split}
	\label{eq:stable_set_appendix}
	\end{equation}
	is guaranteed under the mixed policy (\ref{eq:policy_mixed}) for all $s \in \mathcal{C}$. The set $\mathcal{S}_{st}$ contracts as we (a) increase robustness of the control prior (increase $\sigma_m(\gamma_k)$), (b) decrease our dynamic uncertainty/nonlinearity $C_D$, or (c) increase weighting $\lambda$ on the control prior.
\end{theorem}

	\begin{proof}
		~
		
		\textbf{Step (1): } Find a set in which Lemma 2 is satisfied.
		
		Consider the condition in Lemma 2. Since the right hand side is positive (quadratic), we can consider a bound on the stability condition as follows, 		
		\begin{equation}
		\begin{split}
		& | 2 s^T P d(s,a) + \frac{2}{1+\lambda} s^T P B_2 u_e |  < \\
		& ~~~~~~~~~~~~~ s^T (C_1^T C_1 + \frac{1}{\gamma_k^2}PB_1 B_1^T P) s .
		\end{split}
		\label{eq:stability_norm_bound_1}
		\end{equation}
		Clearly any set of $s$ that satisfy condition ($\ref{eq:stability_norm_bound_1}$) also satisfy the condition in Lemma 2. To find such a set, we bound the terms in Condition (\ref{eq:stability_norm_bound_1}) as follows,
		\begin{equation}
		\begin{split} 
		& | 2 s^T P d(s,a) + \frac{2}{1+\lambda} s^T P B_2 u_e |  \\
		& ~~~~~~~~~~~~~~ \leq ~~ 2 | s^T P d(s,a) | + \frac{2}{1+\lambda} | s^T P B_2 u_e |\\
		& ~~~~~~~~~~~~~~ \leq ~~ 2 \| s \|_2 \| P \|_2 C_D + \frac{2}{1+\lambda} \| s \|_2 \| P B_2 \|_2 C_{\pi} ,
		\end{split}
		\label{eq:stability_norm_bound_2}
		\end{equation}
		where the first inequality follows from the triangle inequality; the second inequality uses our bounds on the disturbance, $C_D$ and control input difference $C_{\pi}$, as well as the Cauchy-Schwarz inequality. Now consider the right hand side of Condition (\ref{eq:stability_norm_bound_1}). Recall that  $\sigma_{m} (\gamma_k) = \sigma_{min} (C_1^T C_1 + \frac{1}{\gamma^2_k} P B_1 B_1^T P)$, the minimum singular value. Then the following holds,
		\begin{equation}
		\sigma_m(\gamma_k) \| s \|^2_2 \leq s^T (C_1^T C_1 + \frac{1}{\gamma_k^2}PB_1 B_1^T P) s 
		\label{eq:stability_norm_bound_3}
		\end{equation}
		Using the bounds in (\ref{eq:stability_norm_bound_2}) and (\ref{eq:stability_norm_bound_3}), we can say that Condition (\ref{eq:stability_norm_bound_1}) is guaranteed to be satisfied if the following holds, 
		\begin{equation}
		\begin{split} 
		& 2 \| s \|_2 \| P \|_2 C_D + \frac{2}{1+\lambda} \| s \|_2 \| P B_2 \|_2 C_{\pi} < \sigma_m(\gamma_k) \| s \|^2_2
		\end{split}
		\label{eq:stability_norm_bound_4}
		\end{equation}
		The set for which this condition (\ref{eq:stability_norm_bound_4}) is satisfied can be described by,
		\begin{equation}
		\begin{split}
		&  \mathcal{C} \setminus \mathcal{S}_{st}: \{s \in \mathbb{R}^n: \| s \|_2 > \frac{1}{\sigma_{m} (\gamma_k)} \Big( 2 \| P \|_2 C_D  \\ 
		& ~~~~~~~~~~~~~~~~~~~~~~ + \frac{2}{1+\lambda} \| P B_2 \|_2 C_{\pi} \Big) ~ , ~ s \in \mathcal{C} \}.
		\end{split}
		\label{eq:stable_set_appendix1}
		\end{equation}
		Recall that $\mathcal{C}$ is the set in which the stabilizing $\mathcal{H}^{\infty}$ controller exists. From Lemma 2, $\dot{V}(s) < 0$ for all $s \in \mathcal{C} \setminus \mathcal{S}_{st}$ described by the set (\ref{eq:stable_set_appendix1}).
		
		\textbf{Step (2): } Establish stability and forward invariance of $\mathcal{S}_{st}$.
		
		The Lyapunov function $V(s) = s^T P s$ decreases towards the origin, and we have established that the time derivative of the Lyapunov function is negative for $s$ in set (\ref{eq:stable_set_appendix1}).
		Therefore, any state $s$ described by the set (\ref{eq:stable_set_appendix1}) (intersected with $\mathcal{C}$) must move towards the origin (i.e. towards $\mathcal{S}_{st}$). This follows directly from the properties of Lyapunov functions. Therefore, the set $\mathcal{S}_{st}$ described in (\ref{eq:stable_set_appendix}) must be asymptotically stable and forward invariant for all $s \in \mathcal{C}$.
		
	\end{proof}

\section{Description of Experiments}

\subsection{Experimental Car-Following}

In the original car-following experiments, a chain of 8 cars followed each other on an 8-mile segment of a single-lane public road. We obtain position (via GPS), velocity, and acceleration data from each of the cars. We cut this data into 4 sets of chains of 5 cars, in order to maximize the data available to learn from. We then cut this into 10 second ``episodes'' (100 data points each). We shuffle these training episodes randomly before each run and feed them to the algorithm, which learns the controller for the $4^{th}$ car in the chain. 

The reward function we use in learning is described below:
\begin{equation}
\begin{split}
    & r = - \dot{v} \min(0,a) - 100 | G_1(s) | - 50 G_2(s) , \\
    & ~ G_1(s) = \begin{cases}
\ \frac{1}{s_{front} - s_{curr}} &\text{if $s_{front} - s_{curr} \leq 10$}\\
\frac{1}{s_{curr} - s_{back}} &\text{if $s_{curr} - s_{back} \leq 10$} \\
0 &\text{otherwise} \\
\end{cases} \\
& ~ G_2(s) = \begin{cases}
\ 1 &\text{if $s_{front} - s_{curr} \leq 2$} \\
1 & \text{if $s_{curr} - s_{back} \leq 2$} \\
0 &\text{otherwise} \\
\end{cases}
\end{split}
\end{equation}
where $s_{curr}$, $s_{front}$, and $s_{back}$ denote the position of the controlled car, the car in front of it, and the car behind it. Also, $a$ denotes the control action (i.e. acceleration/deceleration), and $\dot{v}$ denotes the velocity of the controlled car. Therefore, the first term represents the fuel efficiency of the controlled car, and the other terms encourage the car to maintain headway from the other cars and avoid collision.

The control prior we utilize is a simple bang-bang controller that (inefficiently) tries to keep us between the car and front and back. It is described by,
\begin{equation}
\begin{split}
    & a = \begin{cases}
\ 2.5 &\text{if $K_p \Delta s + K_d \Delta v > 0$}\\
-5 &\text{if $K_p \Delta s + K_d \Delta v < 0$} \\
0 &\text{otherwise} \\
\end{cases} \\
& \Delta s = s_{front} - 2 s_{curr} - s_{back} \\
& \Delta v = v_{front} - 2 v_{curr} - v_{back}
\end{split}
\end{equation}
where $v_{curr}$, $v_{front}$, and $v_{back}$ denote the velocity of the controlled car, the car in front of it, and the car behind it. We set the constants $K_p = 0.4$ and $K_d = 0.5$. Essentially, the control prior tries to maximize the distance from the car in front and behind, taking into account velocities as well as positions.

\subsection{TORCS Racecar Simulator}

In its full generality \textsc{Torcs} provides a rich environment with input from up to $89$ sensors, and optionally the 3D graphic from a chosen camera angle in the race. The controllers have to decide the values of up to $5$ parameters during game play, which correspond to the acceleration, brake, clutch, gear and steering of the car. Apart from the immediate challenge of driving the car on the track, controllers also have to make race-level strategy decisions, like making pit-stops for fuel. A lower level of complexity is provided in the \emph{Practice Mode} setting of \textsc{Torcs}. In this mode all race-level strategies are removed. Currently, so far as we know, state-of-the-art DRL models are capable of racing only in Practice Mode, and this is also the environment that we use. In this mode we consider the input from $29$ sensors, and decide values for the acceleration, steering and brake actions.

The control prior we utilize is a linear controller of the form:
\begin{equation}
    K_p(\epsilon - o_i) + K_i \sum_{j=i-N}^i (\epsilon - o_j) + K_d(o_{i-1} - o_i)
\end{equation}
Where $o_i$ is the most recent observation provided by the simulator for a chosen sensor, and $N$ is a predetermined constant. We have one controller for each of the actions, acceleation, steering and braking.

The pseudo-reward used during training is given by:
\begin{equation}
    r_t = V \cos(\theta) -V \sin(\theta) - V |\texttt{trackPos}|
\end{equation}
Here $V$ is the velocity of the car, $\theta$ is the angle the car makes with the track axis, and \texttt{trackPos} provides the position on the track relative to the track's center. This reward captures the aim of maximizing the longitudinal velocity, minimizing the transverse velocity, and penalizing the agent if it deviates significantly from the center of the track.

\subsection{CartPole Stabilization}

The CartPole simulator is implemented in the OpenAI gym environment ('CartPole-v1'). The dynamics are the same as in the default, as described below,
\begin{equation}
\begin{split}
& \theta_{t+1} = x_t + \dot{x}\tau , \\
& \dot{\theta}_{t+1} = \dot{\theta}_{t}  + \Big( \frac{M g \sin{\theta} - F \cos{\theta} - m l \dot{\theta}^2 \sin{\theta} \cos{\theta}}{\frac{4}{3}M l - m l \cos^2 \theta} \Big) \tau , \\
& x_{t+1} = x_t + \dot{x}\tau , \\
& \dot{x}_{t+1} =  \dot{x}_{t} + \Big( \frac{F + m l \dot{\theta}^2 \sin{\theta} - m l \ddot{\theta} \cos{\theta}}{M} \Big) \tau , 
\end{split}
\end{equation}
where the only modification we make is that the force on the cart can take on a continuous value, $F \in [-10,10]$, rather than 2 discrete values, making the action space much larger. Since the control prior can already stabilize the CartPole, we also modify the reward to characterize \textit{how well} the control stabilizes the pendulum. The reward function is stated below, and incentivizes the CartPole to keep the pole upright while minimizing movement in the x-direction:
\begin{equation}
    r = -100 | \theta | - 2 x^2.
\end{equation}

\section{Control Theoretic Stability Guarantees}

This section in the Appendix goes over the same material in Section 5, but goes into more detail on the $\mathcal{H}^{\infty}$ problem definition. Consider the linear dynamical system described by:
\begin{equation}
\begin{split}
\dot{s} & = A s + B_1 w + B_2 a \\
z & = C_1 s + D_{11} w + D_{12} a \\
y & = C_2 s + D_{21} w + D_{22} a 
\end{split}
\label{eq:dynamics_linear_appendix}
\end{equation}
\noindent
where $w \in \mathbb{R}^{m_1}$ is the disturbance vector, $u \in \mathbb{R}^{m_1}$ is the control input vector, $z \in \mathbb{R}^{p_1}$ is the error vector (controlled output), $y \in \mathbb{R}^{p_2}$ is the observation vector, and $s \in \mathbb{R}^n$ is the state vector. The system transfer function is denoted,
\begin{equation}
\begin{split}
P^{s}(s) &= 
\left(\begin{matrix}
P^{s}_{11} & P^{s}_{12} \\
P^{s}_{21} & P^{s}_{22}
\end{matrix} \right) \\
&=
\left(
\begin{matrix}
D_{11} & D_{12} \\
D_{21} & D_{22} \\
\end{matrix} \right)
+
\left[
\begin{matrix}
C_1 \\
C_2
\end{matrix}
\right]
\left( s I - A \right)^{-1}
\left[
\begin{matrix}
B_1 & B_2
\end{matrix}
\right] , 
\end{split}
\end{equation}
\noindent
where $A, B_i, C_i, D_{ij}$ are defined by the system model (\ref{eq:dynamics_linear_appendix}). Let us make the following assumptions,
\begin{itemize}
	\item The pairs $(A, B_2)$ and $(C_2, A)$ are stabilizable and observable, respectively.
	\item The algebraic Riccati equation $A^T P + PA + C_1^T C_1 + P (B_2 B_2^T - \frac{1}{\gamma_k^2} B_1 B_1^T) P = 0$ has positive-semidefinite solution $P$,
	\item The algebraic Riccati equation $A P_Y + P_Y A^T + B_1^T B_1 =  P_Y (C_2 C_2^T - \frac{1}{\gamma_k^2} C_1 C_1^T) P_Y$ has positive-semidefinite solution $P_Y$,
	\item The matrix $\gamma I - P_Y P$ is positive definite.
\end{itemize}
Under these assumptions, we are guaranteed existence of a stabilizing linear $\mathcal{H}^{\infty}$ controller, $u^{\mathcal{H}^{\infty}} = -K s$ \cite{Doyle1989}. The closed-loop transfer function from disturbance, $w$, to controlled output, $z$, is:
\begin{equation}
T_{wz} = P^{s}_{11} + P^{s}_{12}K(I - P^{s}_{22}K)^{-1}P^{s}_{21}. 
\end{equation}
\noindent
Let $\sigma(\cdot)$ denotes the maximum singular value of the argument, and recall that $\| T_{wz} \|_{\infty} := \sup_w \sigma(T_{wz}(jw))$. Then the $H_{\infty}$ controller solves the problem,
\begin{equation}
\begin{split}
\min_K \sup_w \sigma(T_{wz}(jw)) = \gamma_k,
\end{split}
\end{equation}
\noindent
to give us controller $u^{\mathcal{H}^{\infty}} = -K s$. This generates the maximally robust controller so that the \textit{worst-case disturbance} is attenuated by factor $\gamma_k$ in the system before entering the controlled output. We can synthesize the $H_{\infty}$ controller using techniques described in \cite{Doyle1989}.

The $H_{\infty}$ controller is defined as $u^{\mathcal{H}_\infty} = - B_2^T P x$, where $P$ is a positive symmetric matrix satisfying the Algebraic Riccati equation,
\begin{equation}
A^T P + PA + C_1^T C_1 + \frac{1}{\gamma_k^2} P B_1 B_1^T P - P B_2 B_2^T P = 0,
\label{eq:riccatti_appendix}
\end{equation}
\noindent
where ($A, B_1, B_2, C_1$) are defined in (\ref{eq:dynamics_linear_appendix}). The result is that the control law $u^{\mathcal{H}_\infty}$ stabilizes the system with disturbance attenuation $\| T_{wz} \|_{\infty} \leq \gamma_k$.

Since we are not dealing with a linear system, we need to consider a modification to the dynamics (\ref{eq:dynamics_linear_appendix}) that \textit{linearizes} the dynamics about some equilibrium point and gathers together all non-linearities and disturbances,
\begin{equation}
\dot{s} = f_c(s,a) = A s + B_2 a + d(s,a),
\label{eq:dynamics_linear_uncertain_appendix}
\end{equation}
\noindent
where $d(s,a)$ captures dynamic uncertainty/nonlinearity as well as disturbances. To keep this small, we could use feedback linearization based on our nominal nonlinear model (\ref{eq:dynamics_affine}), but this is outside the scope of this work. 

Consider the Lyapunov function $V(s) = s^T P s$, where P is taken from Equation (\ref{eq:riccatti_appendix}). We can analyze stability of the uncertain system (\ref{eq:dynamics_linear_uncertain}) under the mixed policy (\ref{eq:policy_mixed}) using Lyapunov analysis. We can utilize Lemma 2 in this analysis (see Appendix C) in order to compute a set $\mathcal{S}_{st}$ such that $\dot{V}(s) < 0$ in a region outside that set. Satisfaction of this condition guarantees forward invariance of that set \cite{Khalil2000}, as well as its asymptotic stability (from the region for which $\dot{V}(s) < 0$). 

By bounding terms as described in Section 5, we can conservatively compute the set $\mathcal{S}_{st}$ for which $\dot{V}(s) < 0$, which is shown in Theorem 2. See Appendix D for the derivation of the set (i.e. proof of Theorem 2).

\section{PPO + TRPO Results}

\begin{figure*}[!h]
	\centering
	\includegraphics[scale=0.61]{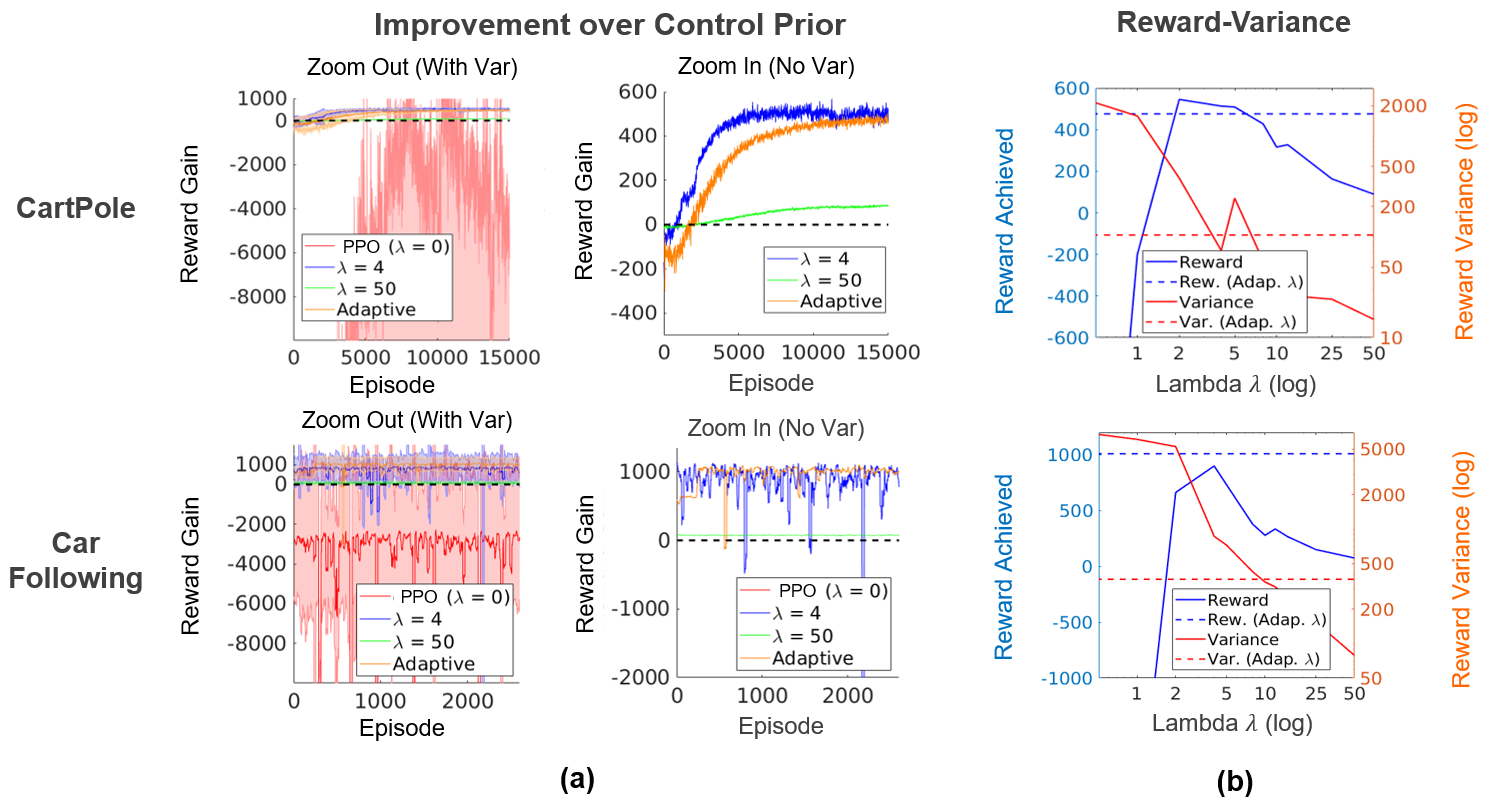}
	\caption{Learning results for CartPole and Car-Following Problems using PPO. (a) Reward improvement over control prior with different set values for $\lambda$ or an adaptive $\lambda$. The right plot is a zoomed-in version of the left plot without variance bars for clarity. Values above the dashed black line signify improvements over the control prior. (b) Performance and variance in the reward as a function of the regularization $\lambda$, across different runs of the algorithm using random initializations/seeds. Dashed lines show the performance (i.e. reward) and variance using the adaptive weighting strategy. Variance is measured for all episodes across all runs. Again, performance is baselined to the control prior, so any performance value above 0 denotes improvement over the control prior.}
	\label{fig:PPO_appendix}
\end{figure*}

\begin{figure*}[!h]
	\centering
	\includegraphics[scale=0.61]{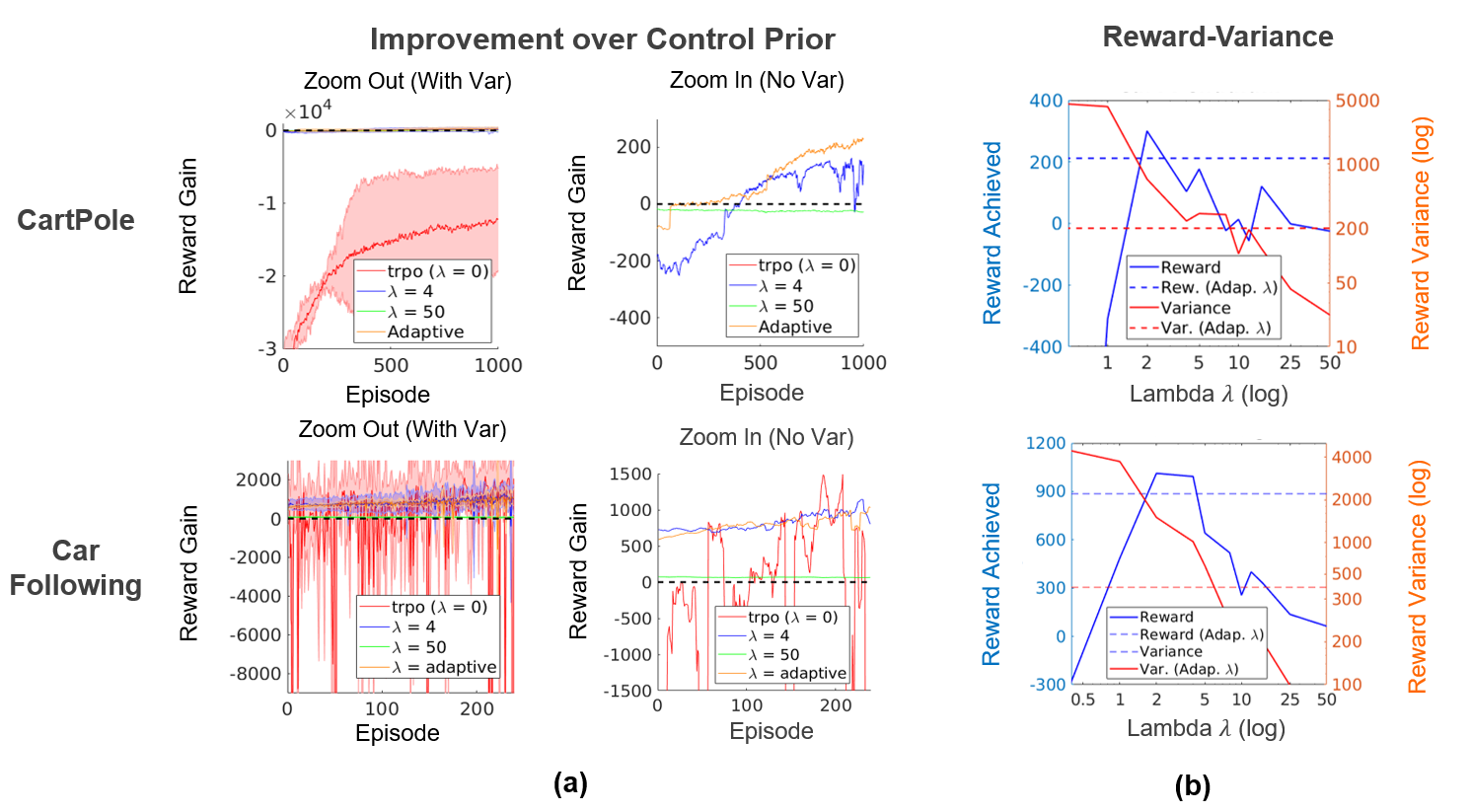}
	\caption{Learning results for CartPole and Car-Following Problems using TRPO. (a) Reward improvement over control prior with different set values for $\lambda$ or an adaptive $\lambda$. The right plot is a zoomed-in version of the left plot without variance bars for clarity. Values above the dashed black line signify improvements over the control prior. (b) Performance and variance in the reward as a function of the regularization $\lambda$, across different runs of the algorithm using random initializations/seeds. Dashed lines show the performance (i.e. reward) and variance using the adaptive weighting strategy. Variance is measured for all episodes across all runs. Again, performance is baselined to the control prior, so any performance value above 0 denotes improvement over the control prior.}
	\label{fig:TRPO_appendix}
\end{figure*}

We also ran all experiments using Proximal Policy Optimization (PPO) or Trust Region Policy Optimization (TRPO) in place of DDPG. The results are shown in Figures \ref{fig:PPO_appendix} and \ref{fig:TRPO_appendix}. The trends mirror those seen in the main paper using DDPG. Low values of $\lambda$ exhibit significant deterioration of performance, because of the larger policy search space. High values of $\lambda$ also exhibit lower performance because they heavily constrain learning. Intermediate $\lambda$ allow for the best learning, with good performance and low variance. Furthermore, adaptive strategies for setting $\lambda$ allows us to better tune the reward-variance tradeoff.

Note that we do not show results for the TORCS Racecar. This is because we were not able to get the baseline PPO or TRPO agent to complete a lap throughout learning. The code for the PPO, TRPO, and DDPG agent for each environment can be found at \url{https://github.com/rcheng805/CORE-RL}.

\end{document}